\documentclass[acmlarge]{acmart}

\AtBeginDocument{%
  }

\usepackage{lineno}
\usepackage{multirow}
\usepackage{siunitx}
\usepackage{booktabs}
\usepackage{adjustbox}
\usepackage{tabularx}
\usepackage{colortbl}
\usepackage{xcolor}
\usepackage[normalem]{ulem}
\usepackage{makecell}
\usepackage{pifont}
\usepackage[linesnumbered,ruled,vlined]{algorithm2e}
\usepackage{ulem}
\usepackage{soul}

\usepackage{color}
\usepackage{listings}
\usepackage{subcaption}
\usepackage{float}
\usepackage{xspace}
\usepackage[most]{tcolorbox}
\newcommand{\workname}{DeepFeature\xspace}
\newcommand{\modify}[1]{{\color{black}#1}} 
\newcommand{\final}[1]{{\color{black}#1}}

\newtcolorbox{promptbox}[1][]{
    breakable,
    enhanced,
    colback=gray!5,
    colframe=gray!70,
    fonttitle=\bfseries,
    fontupper=\small\ttfamily,
    before upper={\setlength{\parskip}{0.4em}},
    title={#1}
}

\setcopyright{cc}
\setcctype{by}
\acmJournal{IMWUT}
\acmYear{2026} \acmVolume{10} \acmNumber{3} \acmArticle{131}
\acmMonth{9} \acmDOI{10.1145/3831996}

\makeatletter
\renewcommand\@fnsymbol[1]{%
  \ifcase#1\or $\dagger$\or $*$\or $\ddagger$\or $\S$\or $\P$%
  \else\@ctrerr\fi}
\makeatother

\begin{document}

\title{\workname: LLM-Empowered Context-aware Feature Generation for Wearable Biosignals}

\author{Kaiwei Liu}
\authornote{Co-author equal contribution.}
\orcid{0009-0002-4108-0898}
\affiliation{
  \institution{The Chinese University of Hong Kong}
  \city{Hong Kong}
  \country{China}}
\email{kaiweiliu@link.cuhk.edu.hk}

\author{Yuting He}
\authornotemark[1]
\orcid{0009-0003-1954-952X}
\affiliation{
  \institution{The Chinese University of Hong Kong}
  \city{Hong Kong}
  \country{China}}
\email{heyuting@link.cuhk.edu.hk}

\author{Bufang Yang}
\orcid{0000-0003-0032-2539}
\affiliation{
  \institution{The Chinese University of Hong Kong}
  \city{Hong Kong}
  \country{China}}
\email{bfyang@link.cuhk.edu.hk}

\author{Mu Yuan}
\orcid{0000-0002-2624-8755}
\affiliation{
  \institution{The Chinese University of Hong Kong}
  \city{Hong Kong}
  \country{China}}
\email{ym0813@mail.ustc.edu.cn}

\author{Chun Man Victor Wong}
\orcid{0000-0002-4180-1459}
\affiliation{
  \institution{The Education University of Hong Kong}
  \city{Hong Kong}
  \country{China}}
\email{s1138617@s.eduhk.hk}

\author{Ho Pong Andrew Sze}
\orcid{0009-0009-5268-9624}
\affiliation{
  \institution{Hong Kong Applied Science and Technology Research Institute}
  \city{Hong Kong}
  \country{China}}
\email{sze@astri.org}

\author{Guoliang Xing}
\orcid{0000-0003-1772-7751}
\affiliation{
  \institution{The Chinese University of Hong Kong}
  \city{Hong Kong}
  \country{China}}
\email{glxing@cuhk.edu.hk}

\author{Zhenyu Yan}
\orcid{0000-0002-4433-5211}
\affiliation{
  \institution{The Chinese University of Hong Kong}
  \city{Hong Kong}
  \country{China}}
\email{zyyan@ie.cuhk.edu.hk}

\author{Hongkai Chen}
\authornote{Corresponding author.}
\orcid{0000-0001-7206-6584}
\correspondingauthor
\affiliation{
  \institution{The Chinese University of Hong Kong}
  \city{Hong Kong}
  \country{China}}
\email{hkchen@ie.cuhk.edu.hk}

\begin{abstract}
Biosignals collected from wearable devices are widely utilized in healthcare applications. Machine learning models used in these applications often rely on features extracted from biosignals due to their effectiveness, lower data dimensionality, and wide compatibility across various model architectures. However, existing feature extraction methods often lack task-specific contextual knowledge, struggle to identify optimal features in high-dimensional combinatorial feature space, and are prone to automated code generation and execution errors. In this paper, we propose \workname, the first LLM-empowered, context-aware feature generation framework for wearable biosignals. \workname introduces a multi-source feature generation mechanism that integrates the inherent ability of LLMs, expert knowledge and inter-feature interactions. It also employs an iterative feature refinement process that uses feature assessment-based feedback for feature re-selection. Additionally, \workname utilizes a robust multi-layer filtering and verification approach for feature description-to-code translation to ensure that the feature extraction functions run without crashing. \final{Experimental evaluation results show that \workname achieves the highest average AUROC across eight tasks under both sample-level and subject-level settings, outperforming the best baselines by 4.60\% and 4.61\%, respectively. \workname achieves the most pronounced gains on the PPG-BP tasks, while remaining competitive with the best-performing baselines on Epilepsy, WESAD, and our self-collected SEN dataset.}

\end{abstract}

\makeatletter
\def\@correspondingauthormark{}
\def\@mkauthorsaddresses{%
  \ifnum\num@authors>1\relax
  Authors' \else Author's \fi
  Contact Information:
  \bgroup
  \def\streetaddress##1{\unskip\ignorespaces}%
  \def\postcode##1{\unskip\ignorespaces}%
  \def\position##1{\unskip\ignorespaces}%
  \gdef\@ACM@institution@separator{, }%
  \def\institution##1{\unskip\@ACM@institution@separator
    ##1\gdef\@ACM@institution@separator{ and }}%
  \def\city##1{\unskip, ##1}%
  \def\state##1{\unskip, ##1}%
  \renewcommand\department[2][0]{\unskip\@addpunct, ##2}%
  \def\country##1{\unskip, ##1}%
  \def\and{\unskip; \gdef\@ACM@institution@separator{, }}%
  \def\@author##1{##1}%
  \def\@correspondingauthormark{\unskip\ (corresponding author)}%
  \def\email##1##2{\unskip, ##2}
  \addresses
  \egroup}
\makeatother




\maketitle

\renewcommand{\shortauthors}{Liu et al.}
\section{INTRODUCTION} \label{intro}

\begin{figure}
  \centering
  \includegraphics[width=\linewidth]{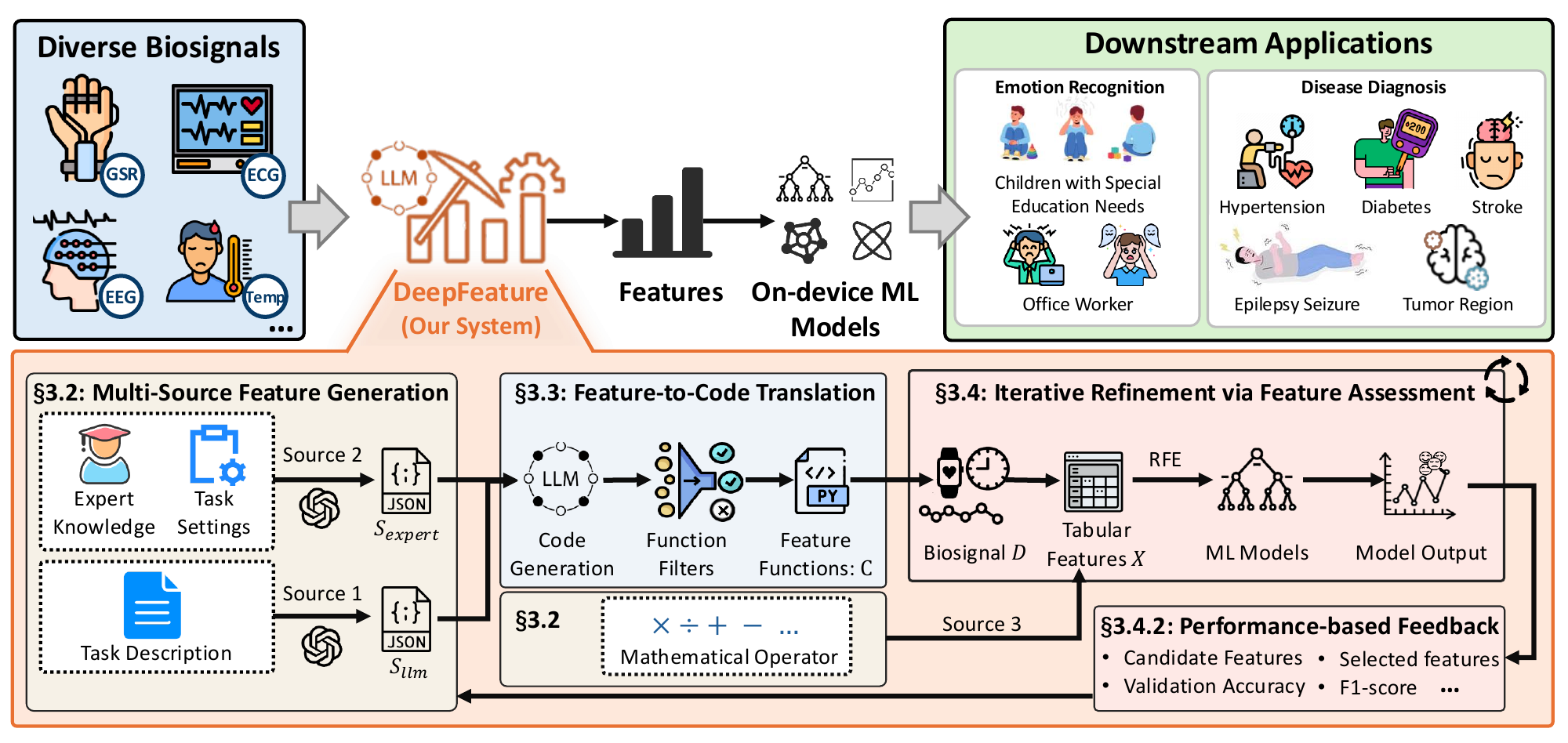}
  \caption{
  \modify{
  The \workname framework for automated, context-aware feature generation. While training ML models using raw biosignals often yields suboptimal performance, \workname automates feature engineering from multi-modal data (e.g., GSR, ECG, EEG). As shown in the lower panel, the system leverages LLMs to synthesize feature descriptions by integrating task settings with expert knowledge. These features are translated into executable code to transform raw signals into structured tabular data. After enhancement with mathematical operators, the data is used to train downstream models, with performance metrics feeding back into the LLM to iteratively refine the feature set.}
  } 
  \label{fig:overview}
\end{figure}

Wearable devices, such as smartwatches, rings, chest straps, and headbands equipped with multimodal physiological sensors, continuously capture time-series biosignals, such as electrocardiogram (ECG), galvanic skin response (GSR), and electroencephalogram (EEG).
According to 
Grand View Research~\cite{grandview2024china}, the Chinese wearable device market exhibits strong growth potential, with projections indicating a 28.5\% compound annual growth rate (CAGR) from 2024 to 2030.
As shown in Figure~\ref{fig:overview}, these devices are widely utilized in numerous healthcare applications, such as emotion recognition for diverse populations~\cite{choksi2024sensemo,schmidt2018introducing, healey2005detecting, cano2024wearable}, disease diagnosis~\cite{he2025opentcm,yang2024drhouse}, and monitoring and managing chronic conditions (e.g., cardiovascular diseases, epilepsy, and diabetes)~\cite{li2024emomarker,chen2019committed,paoletti2019data,adepoju2024wearable, liang2018new, andrzejak2001indications,qi2025alzheimer}.
However, training machine learning (ML) models to achieve desirable generalizability using raw biosignals remains challenging~\cite{galaz2022comparison, bento2022comparing, lin2020comparison, gallo2025towards, cote2020interpreting}. 
In particular, using raw multimodal biosignals to train deep learning models is hindered by factors including high data volume demand, privacy concerns, and regulatory constraints, making it a suboptimal approach~\cite{li2024survey}.
In contrast, feature-based solutions have proven to be more effective for handling wearable biosignals with several advantages.
First of all, 
% handcrafted 
they can effectively capture key information from biosignals in \textit{relatively small-scale datasets}, enabling superior few-shot learning performance~\cite{lin2020comparison} and improved domain generalization~\cite{bento2022comparing} across diverse healthcare applications.
Additionally, these methods are well-suited for \textit{local deployment and real-time inference} on resource-constrained wearable devices due to lower data dimensionality and model complexity~\cite{shukla2020half,georgiou2021comparison}.
Furthermore, feature-based solutions are \textit{compatible with a wide range of lightweight ML models} (e.g., SVM, Random Forest, XGBoost)~\cite{shoeibi2021comprehensive}, 
providing deployment flexibility across various hardware platforms.
For these reasons, feature extraction remains a critical approach for enabling wearable biosignal-based healthcare applications.

However, existing feature extraction methods have significant limitations when applied to wearable biosignals.
% First
First, existing feature extraction methods either rely on predefined feature sets~\cite{christ2018time, barandas2020tsfel, Kats2021, lubba2019catch22} or utilize off-the-shelf Large Language Models (LLMs) with general task description for feature generation~\cite{gong2025evolutionary, hollmann2024large, nam2024optimized, han2024large}.
Nonetheless, healthcare tasks based on wearable biosignals are highly context-specific, encompassing unique task settings (e.g., data collection protocols, sensor modalities) and task-relevant expert knowledge (e.g., physiological characteristics of target groups, feature selection priors).
In addition, biosignals often exhibit inter- and intra-individual variability due to physiological and psychological factors.
This context-specific nature makes the rapid identification of an effective, customized feature set in such applications particularly challenging.
%Second
\final{Second, wearable biosignal applications typically involve multiple physiological signals, each with a range of extractable features and feature-specific parameter configurations. Moreover, features can be combined across signals to form composite features. These factors result in an extremely high-dimensional combinatorial feature space.}
\final{Navigating such a large feature space to select optimal feature set is highly error-prone, frequently leading to suboptimal model performance.}
Existing solutions, however, either fail to \final{effectively} assess the quality of \final{the extracted features}~\cite{jeong2024llm, li2025exploring, han2024large} or merely rely on iterative interactions with LLMs using invariant prompts, with the expectation of achieving improved outcomes~\cite{hollmann2024large, gong2025evolutionary}. 
% Last
Last, the vast number and heterogeneity of applications and extracted features make it impractical to manually implement the feature extraction functions. 
Recent work~\cite{shen2025autoiot} has explored LLM-based automated approaches for implementing and executing the feature extraction functions from sensor data.
However, we have observed that it often generates erroneous feature extraction functions in the context of wearable biosignals, and its code improvement module frequently exceeds the token limits of the LLM (see Section~\ref{subsec:manualeffort}). 
This emphasizes the demand for \final{a fine-grained and verifiable mechanism that can automatically and correctly translate feature extraction processes into executable code.}

\final{Our goal is to automatically generate and extract effective, customized feature sets for a given wearable biosignal-based application, enabling high-performing downstream models.}
To address the above limitations, in this paper, we propose \textit{\workname}, the first LLM-empowered, context-aware feature generation framework specifically designed for wearable biosignals.
First, we introduce a \modify{multi-source} feature generation approach 
that integrates the inherent knowledge of LLMs, expert knowledge from well-established literature libraries, and inter-feature interactions.
\final{The three feature sources are entirely distinct yet highly complementary, integrating comprehensive task contexts to support the identification of application-specific features.}
Additionally, we design a ranking-based feature elimination module and a model performance-based feedback construction method to guide the LLM in improving feature selection.
\final{\workname constructs structured and instructive feedback for the LLM, which summarizes downstream model performance at global, class, and class-pair levels, ranks weak classes and class pairs to expose key performance bottlenecks.}
\final{It guides the LLM in refining feature-set selection within the large feature space.}
Finally, to ensure verifiable feature generation and automated extraction, we propose a multi-layer filtering and verification approach for feature extraction functions that enhances the robustness of feature description-to-code translation, \final{which is designed based on the practical failure modes and key constraints of wearable biosignal feature extraction.}
To evaluate \workname, we utilize three publicly available wearable biosignal datasets encompassing seven healthcare tasks.
To further diversify the evaluation dataset and cover a broader range of application scenarios, we recruit 29 students with special education needs (SEN) and collect 304 MB of multimodal biosignal data using wristbands over a total duration of approximately 44 hours. 
During data collection, licensed teachers were asked to label students’ emotions for emotion recognition tasks.
% \modify{Our evaluation shows that, under the sample-level split, \workname achieves the highest average AUROC of 80.62\%, outperforming the best baseline by 4.56\% and achieving improvements of up to 9.37\% on individual tasks. Under the more rigorous subject-level split, \workname also achieves the highest average AUROC, outperforming the best baseline by 4.61\%, demonstrating strong generalization to unseen users.}
\final{
Our evaluation shows that, under the sample-level split, \workname achieves the highest average AUROC across eight tasks, outperforming the best baseline by 4.60\%. 
Under the more rigorous subject-level split, \workname also achieves the highest average AUROC, outperforming the best baseline by 4.61\%. 
The improvements are most pronounced on the PPG-BP tasks, while \workname remains competitive with the best-performing baselines on Epilepsy, WESAD, and SEN.
}
% Additionally, we observe notable variation in optimal features across tasks; for example, emotion recognition in typical individuals relies on autonomic nervous system signals (e.g., GSR), whereas SEN students rely more on movement patterns (e.g., acceleration).
\final{Additionally, we observe notable variation in selected features across tasks; for example, WESAD emotion recognition emphasizes autonomic nervous system signals such as GSR, whereas the selected SEN features place greater weight on movement-related signals such as acceleration.}

Our main contributions are summarized as follows.
\begin{itemize}
    \item We propose \workname, the first LLM-empowered, context-aware feature generation framework specifically designed for wearable biosignals. 
    The generated features are effective with satisfactory performance in wearable biosignal-based healthcare applications, enabling local deployment and real-time inference.
    \item We propose a multi-source feature generation mechanism that integrates the inherent knowledge of LLMs, task-specific context, and inter-feature interactions. 
    This approach provides comprehensive task-specific context to LLMs, enabling tailored feature generation.
    \item We develop an iterative feature generation strategy with in-loop feedback, enabling LLMs to adaptively refine the generated feature set over time. 
    Furthermore, we introduce a feature description-to-code translation pipeline that incorporates function filters and an output verification module, ensuring the quality of both feature extraction code generation and its execution.
    % \item We extensively evaluate \workname on one self-collected and 3 publicly available datasets covering 8 healthcare tasks.  
    % \modify{The experimental results demonstrate that \workname achieves the highest average AUROC under both sample-level and subject-level evaluation settings, with improvements of 4.56\% and 4.61\% over the best baselines, respectively.}
    \item \final{
    We extensively evaluate \workname on one self-collected and three publicly available datasets covering eight healthcare tasks. The experimental results demonstrate that \workname achieves the highest average AUROC under both sample-level and subject-level settings, with particularly pronounced gains on PPG-BP tasks and competitive performance on Epilepsy, WESAD, and SEN tasks.
    }
\end{itemize}

\section{MOTIVATION STUDY}

ML models for healthcare wearables typically process either raw biosignals or extracted features.
Although raw biosignal-based approaches have shown promise, they often suffer from high training data demand~\cite{li2024survey}, limited  generalizability~\cite{galaz2022comparison, bento2022comparing, gallo2025towards} and significant inference overhead~\cite{georgiou2021comparison}.
In contrast, feature-based approaches~\cite{zhang2022predicting, park2024hide} are generally more effective and flexible for resource-constrained wearable devices. 

However, existing feature extraction methods have several limitations when applied to wearable biosignals: 1) Highly specialized task context makes the identification of effective features particularly challenging. 2) High-dimensional combinatorial feature space caused by complex and diverse multimodal biosignals makes the determination of optimal feature combination difficult. 3) The vast number and complexity
of features to be extracted makes it impractical to manually implement the extraction functions. 
In this section, we motivate our study using these challenges and explore the opportunities to address them.

% \begin{figure}[t]
%     \centering
%     \includegraphics[width=0.8\linewidth]{fig/expert_motivation.pdf}
%     \caption{Performance comparison of features directly generated by LLMs (with and without feedback) and expert-validated features on WESAD dataset.}
%     \label{fig:expert}
% \end{figure}

\begin{figure}[t]
    \centering
    \begin{minipage}[t]{0.38\textwidth}
        \centering
        \includegraphics[width=\linewidth]{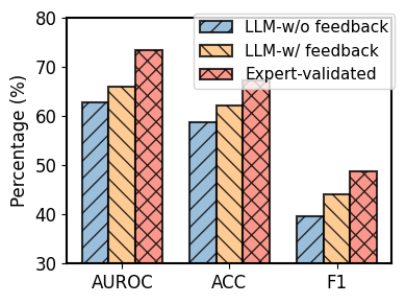}
        \caption{Performance comparison of features directly generated by LLMs (with and without feedback) and expert-validated features on WESAD dataset.}
        \label{fig:expert}
    \end{minipage}
    \hfill
    \begin{minipage}[t]{0.51\textwidth}
        \centering
        \includegraphics[width=\linewidth]{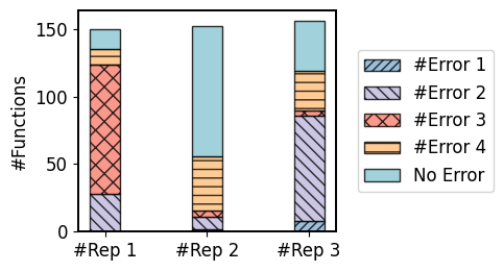}
        \caption{Error distribution in code generation and execution across three independent trials (Repetition 1--3). The errors are categorized as follows: Error 1 (import failures), Error 2 (runtime crashes), Error 3 (incomplete implementations), and Error 4 (logical flaws).}
        \label{fig:errors}
    \end{minipage}
\end{figure}

\subsection{Context-specific Biosignal Features}
Wearable biosignal-based healthcare applications demand specialized contextual knowledge to extract task-tailored features, which is difficult for existing solutions that provide LLMs with general task description~\cite{gong2025evolutionary, hollmann2024large, nam2024optimized, han2024large}.
To understand this limitation, we first conduct preliminary experiments on the WESAD dataset~\cite{schmidt2018introducing} by directly prompting the LLM with general descriptions of the stress and affect detection task\modify{, adapting the prompt template from~\cite{shen2025autoiot}.}
LLM-generated features include basic statistical features (e.g., max/min, mean, standard deviation, skewness).
Expert-validated features~\cite{schmidt2018introducing} incorporate advanced, task-specific features, such as frequency-domain HRV components (e.g., LF/HF ratio) and detailed GSR features (e.g., SCR slopes, startle magnitudes).

Figure~\ref{fig:expert} shows that expert-validated features significantly outperform LLM-generated features across three metrics, including AUROC, accuracy and F1 score, with improvements of 10.64\%, 8.58\%, and 9.09\%, respectively.
% what LLM-generated features we used? SOTA?
It demonstrates that relevant contextual knowledge of the task, such as sensor properties, data collection details, and prior researchers' experience on the task, can potentially influence the constitution of the optimal feature set. 
We also provide LLMs with a literature~\cite{ahmed2019emotion} on emotion recognition system based on body movement, and the LLM draws inspiration and generates a novel feature: \textit{``Difference between chest ACC Y skewness and wrist ACC Z skewness}.'' 
It also provides the rationale: ``\textit{Combines motion asymmetry features from chest and wrist to detect subtle physical responses to emotions}.'' This feature was never generated when prompting LLMs with general task descriptions alone.
This experiment highlights the importance of incorporating task-specific contextual knowledge into the feature generation process.

\modify{
However, we further observe that merely supplying the LLM with corresponding literature causes the generated features to be highly confined to expert knowledge, thereby missing out on simpler, more general, but effective features. This suggests that task-specific contextual knowledge should serve as an important supplement to LLM inherent knowledge, rather than the sole feature source.
}

\subsection{High-dimensional Combinatorial Feature Space}
Wearable biosignals are inherently complex and diverse, creating a high-dimensional combinatorial feature space.
For a single signal modality, this space is defined by the number of feature types and their parameter configurations.
With multi-modal data, the feature space expands exponentially due to the potential interactions and combinations across different modalities.
Manually searching this immense space for optimal feature extraction settings is both error-prone and likely to yield suboptimal model performance.

As shown in Figure~\ref{fig:expert}, to further illustrate this challenge, we conducted an experiment comparing features generated by an LLM in a single attempt against those iteratively refined using model performance feedback (e.g., classification accuracy or feature importance rankings).
On the WESAD dataset, the LLM's initial output consists of basic statistical features. 
After receiving feedback on the trained model's performance, the LLM produces more advanced features in a second iteration, which include the Signal Magnitude Area (SMA) for accelerometer data to capture overall motion intensity and inter-axis correlations to identify movement patterns.
This iterative process significantly improved model performance over a single attempt, with AUROC, accuracy, and F1 scores increasing by 3.23\%, 3.30\%, and 4.43\%, respectively.
These results validate the necessity of an iterative refinement cycle for navigating feature spaces, where downstream performance metrics can guide the progressive optimization of the feature set.

% \begin{figure}[t]
%     \centering
%     \includegraphics[width=0.5\textwidth]{fig/func_errors.pdf} 
%     \caption{Error distribution in code generation and execution across three independent trials (Repetition 1--3). The errors are categorized as follows: Error 1 (import failures), Error 2 (runtime crashes), Error 3 (incomplete implementations), and Error 4 (logical flaws). }
%     \label{fig:errors}
% \end{figure}

% \begin{figure*}[t]
%     \centering
%     \includegraphics[width=\linewidth]{fig/overview.pdf}
%     \captionsetup{skip=1pt}
%     \caption{
%         Overview of \workname. \workname~leverages LLMs to generate feature descriptions by combining task settings with expert knowledge. These features are translated into executable code to transform raw biosignals into tabular data. After being enhanced with mathematical operators, the tabular data is used to train downstream models. The model's performance then feeds back to the LLMs to refine the feature set in the next iteration.
%     }
%     \label{fig:overview}
% \end{figure*}

\subsection{Errors in Feature Extraction Automation}\label{subsec:manualeffort}
Manually implementing feature extraction functions for numerous heterogeneous healthcare applications is impractical.  
While automated feature description-to-code translation offers a solution, its effectiveness is adversely affected by errors during implementation and execution. 
Its iterative process further increases the cumulative probability of such errors.
To better illustrate this issue, we conduct three independent trials in which off-the-shelf LLMs (e.g., DeepSeek-V3~\cite{liu2024deepseek}) are prompted with biosignal-specific feature descriptions to auto-generate and execute extraction function code.
As shown in Figure~\ref{fig:errors}, the experimental results reveal a high prevalence of four common error types in the generated functions.

Recent work~\cite{shen2025autoiot} has explored an error correction mechanism where the execution issue logs are provided to LLMs for code regeneration. 
However, our experiments show that this approach is frequently ineffective for wearable biosignals because the detailed logs almost always exceed the LLM's token limits.
This shows that naively feeding issue logs to LLMs is still impractical in our context, and, consequently, a dedicated mechanism for identifying and eliminating implementation and execution errors is critical.
% \section{System Overview}

% for a given task, \workname~first introduces features from multiple sources (\S\ref{sec:multi_source_feature_generation}). 
% It simultaneously prompts LLMs in two separate sessions: one to directly give new features and another to derive features grounded in expert knowledge. 
% The resulting features from both sessions are merged into the current feature set. All newly introduced features are translated into executable code (\S\ref{sec:code_conversion}) and used to transform the raw biosignal data. The resulting tabular feature data are concatenated with the previous tabular feature data along the feature dimension. 
% Then, \workname~employs a set of predefined operators to generate new features by combining the existing columns in the tabular feature data. These features are then also added to the feature set.
% After that, \workname~utilizes a non-LLM feature selection (\S\ref{sec:feature_selection}) module to select informative columns from the tabular feature data and uses them to train the downstream model.
% After that, the model is evaluated, and the evaluation results (\S\ref{sec:results_analysis}) are used to construct feedback for LLMs as a guidance for multi-source feature generation in the next iteration.

\section{FRAMEWORK DESIGN}\label{system_design}

In this section, we first discuss the overview of \workname, then describe its major modules: \emph{multi-source feature generation} (Section~\ref{sec:multi_source_feature_generation}), \emph{robust feature-to-code translation} (Section~\ref{sec:code_conversion}), and \emph{iterative refinement via feature assessment} (Section~\ref{sec:feature_assessment}).
Finally, we summarize the overall algorithm used in \workname (Section~\ref{sec:overall_algorithm}).

\subsection{Overview}
\workname~is an LLM-empowered, context-aware feature generation framework for wearable biosignals. 
As shown in Figure~\ref{fig:overview}, smart devices equipped with multimodal physiological sensors, such as smartwatches, continuously monitor users and collect time-series biosignals.
At each feature generation iteration, \workname~first generates features (Section~\ref{sec:multi_source_feature_generation}) through direct LLM generation (Source 1) and task-specific context-guided generation (Source 2).
These features are translated into executable code (Section~\ref{sec:code_conversion}), which is then used to transform raw biosignals into tabular feature data. 
After that, \workname~uses mathematical operators to combine columns in the tabular data to introduce new columns, thereby enriching the feature set (Source 3).
\workname~then trains an ML model on the tabular feature data and utilizes the evaluation results of the model to guide the refinement of feature generation in the next iteration (Section~\ref{sec:feature_assessment}). 
This feedback-driven loop ensures continuous improvement in feature quality, improving the performance of downstream ML models.

\subsection{Multi-source Feature Generation}
\label{sec:multi_source_feature_generation}
\begin{figure}[t]
    \centering
    \includegraphics[width=0.85\linewidth]{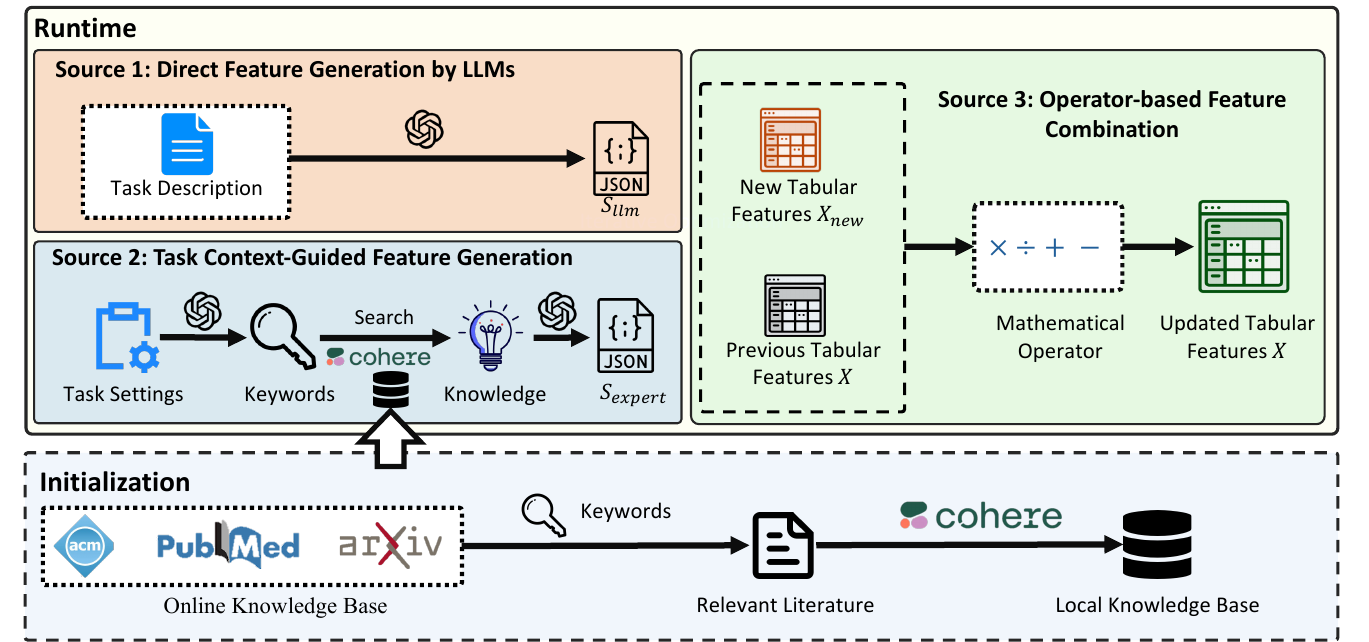}
    \caption{
        Multi-source feature generation. 
    }
    \label{fig:multisource}
\end{figure}

To identify effective features in a task-specific context, we propose a multi-source feature generation mechanism. 
As illustrated in Figure~\ref{fig:multisource}, the initialization stage requires minimal user effort. 
Users only need to provide the following basic information in the prompt, which is typically already available for the given task:

\begin{itemize}
    \item Task Description: Specifies the objectives of the machine learning model, including the classification labels.
    \item Data Collection Protocol: Outlines the procedures, steps, and standardized methods used for data acquisition.
    \item Sensor Modalities: Lists the types of sensors and measurement methods employed (e.g., EEG, GSR).
    \item Subject Characteristics: Describes demographic and clinical details of participants, such as age and health status.
\end{itemize}

\begin{figure}[t]
  \centering
\includegraphics[width=0.85\linewidth]{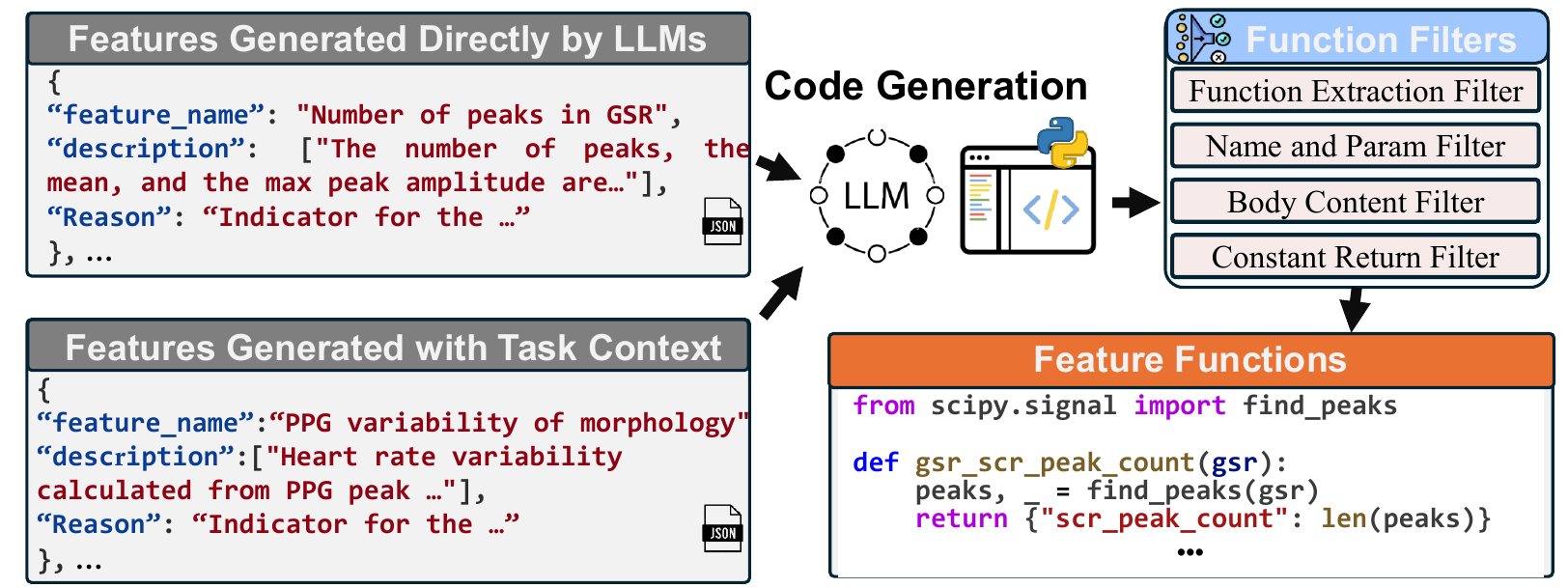}
  \caption{Feature-to-code conversion pipeline. Features in JSON format from two sources are integrated into prompts for LLMs. Then, LLMs generate feature extraction functions, which are then processed by the filters.} 
  \label{fig:filter}
\end{figure}

This information is fed to the LLM for a \textit{single} inference to generate task-relevant keywords.
These keywords are then used to automatically construct the local knowledge base by crawling relevant literature---a process that takes only tens of minutes and requires no manual intervention.
Once initialization is complete, \workname introduces new features from the following three distinct sources:

\noindent\textbf{(Source 1) Direct Feature Generation by LLMs.}  Pretrained on massive corpora, LLMs already contain extensive general knowledge. 
\workname~provides a task description (e.g., \emph{``This task is non-invasive hypertension detection using PPG signal...''}) and directly prompts the LLM to generate features.
The features are returned in a structured JSON format, where each feature includes its name, a brief description, and the rationale for its relevance to the task, as shown in Fig.~\ref{fig:filter}.
\modify{
We retain this source because we empirically find that conditioning LLMs on task-specific context, including task-specific settings and relevant literature, tends to confine the generated features too narrowly to the provided context, causing the model to overlook a broad set of general yet effective features.
}

\noindent\textbf{(Source 2) Task Context-Guided Feature Generation.} 
For its second source, \workname~incorporates feature generation guided by task-specific contextual knowledge, including task settings (e.g., data collection protocols, sensor modalities) and retrieved expert knowledge (e.g., physiological characteristics of target groups, feature selection priors).
The task settings are provided manually during initialization and remain fixed.
The expert knowledge base is built by leveraging three online libraries, namely arXiv~\cite{arxiv}, ACM Digital Library~\cite{acm}, and PubMed~\cite{pubmed}. 
\modify{\workname builds a separate local expert knowledge base for each downstream task, because different tasks involve different sensor modalities, target labels, and physiological contexts.}
\workname first summarizes task-relevant keywords to retrieve massive research literature. 
The retrieved documents are split into smaller text chunks, processed into text embeddings, and stored locally. 
At runtime, \workname queries LLMs to generate a set of task-related keywords, which are then converted into a text embedding to search the knowledge base for the most relevant text chunks. 
These chunks, alongside the task settings, are then integrated into the prompts for feature-generation.

\noindent\textbf{(Source 3) Operator-based Feature Combination.} 
We notice that the features generated from the first two sources usually rely on a single biosignal modality, lacking the interaction between different modalities.
To address this issue, we incorporate the third source involving mathematical operators between different features.
\modify{
Some studies~\cite{qian2023limitations, deng2024language} reveal that LLMs are generally better at semantic understanding and textual reasoning, but less reliable for systematic numerical exploration and enumeration.
Therefore, rather than relying on LLMs for this source as prior works do~\cite{hollmann2024large}, we adopt a non-LLM-based method that systematically enumerates combinations of existing features using the prescribed operators, decoupling feature combination from semantic reasoning. 
}
\modify{
In this work, we use four arithmetic operators in source 3: division ``$/$'', multiplication ``$\times$'', subtraction ``$-$'', and addition ``$+$'', which is empirically validated by existing works~\cite{hollmann2024large, gong2025evolutionary}.
\workname incorporates unary operators (e.g., $\sqrt{}$, $\sin$, $\cos$) appeared in these works through the feature functions provided by source 1 and source 2.
Furthermore, \workname can also introduce more complex operators by referring expert knowledge in source 2.
}
The details of source 3 are as follows.
After translating the features from the first two sources into code and executing them on the raw biosignal data (see Section~\ref{sec:code_conversion}), the raw data is transformed into feature values in tabular form. 
Each column represents a feature and each row a data sample.
\workname applies the operators to the original features in this table to produce new composite features. 
Specifically, \workname first computes the mutual information between each column and the label as its importance value. 
Then, \workname selects \textit{k} columns with the largest importance values and pairs them up, leading to $C^2_k$ column pairs. 
Last, \workname applies all operators to each column pair and appends the composite features to the original features.

\workname operates iteratively. 
In each iteration, it aggregates newly generated features from all three sources into a candidate feature set. 
Beyond features explicitly referenced in the literature, \workname frequently produces inspired variations that diverge from established references. 
Additionally, by applying mathematical operators to existing features, \workname creates composite features---many of which are novel and have not previously appeared in the literature.
\modify{
\workname decouples feature generation from code execution. For source 1 and source 2, \workname generates feature descriptions instead of directly generating the executable feature extraction function code. This improves the robustness and scalability of the overall system.
}
The iterative cycle continues until a predefined maximum number of iterations is reached.
Throughout this process, \workname not only expands the feature set but also uncovers valuable new insights that extend beyond current knowledge.

\subsection{Robust Feature-to-Code Translation}
\label{sec:code_conversion}
After obtaining features from source 1 and source 2, \workname adopts robust feature-to-code translation to convert the JSON-formatted textual descriptions of the features from the first two sources into tabular feature data.
This section also introduces our multi-layer function filters and execution verification modules, which reduce execution failures and enforce fixed-dimensional tabular outputs.
\modify{
Unlike existing general-purpose code generation or translation approaches, our design is based on the error distribution observed in practice and tailored to the key constraints of wearable biosignal feature extraction.
}

\subsubsection{Multi-layer Extraction Function Filtering}
\label{sec:filtering}
First, \workname prompts LLMs with the JSON-formatted textual descriptions of features and queries LLMs to translate them into executable code.
To avoid errors in feature extraction function code, we design a mechanism to identify erroneous feature extraction functions before execution. 
Based on the error distribution shown in Figure~\ref{fig:errors}, we develop four filters that detect and remove functions containing import errors and incomplete implementations.  \\
\textbf{Function Extraction Filter.} This filter performs abstract syntax tree (AST) analysis on the code to isolate individual feature extraction function definitions. 
It evaluates the syntactic validity and dependency availability of each function through targeted compilation, incorporating automatic resolution of external dependencies. 
Functions that can be properly compiled and loaded pass this filter, ensuring that only correctly structured implementations can be used for subsequent processes.\\
\textbf{Name and Parameter Filter.} 
\modify{
This filter ensures proper function calls by aligning function names and parameters with biosignals.}
It examines each function's name and parameter structure to ensure alignment with our predefined sensor configurations. 
Functions pass this filter only when their names begin with one or multiple sensor identifiers and accept at least one parameter whose name begins with one or multiple sensor identifiers. \\
\textbf{Body Content Filter.} 
\modify{
This filter addresses cases where LLMs retrieve a feature concept from the literature but fail to convert it into executable logic, as the literature merely mentions the concept without providing the implementation details.
}
It examines function implementations to eliminate placeholder or skeletal functions. 
It checks if a function's body consists solely of a single pass statement, filtering out such functions while retaining functions with substantive code. \\
\textbf{Constant Return Filter.} 
\modify{
This filter identifies cases where LLMs mistakenly treat static reference values from the literature as the feature extraction functions.
}
It analyzes function return patterns to identify and exclude functions that return only constant values or have empty returns. 
By examining the abstract syntax tree of each function, it detects return statements containing literal constants or null returns, and filters out such functions.

\workname integrates these filters sequentially to process the newly generated feature extraction functions, thereby reducing the likelihood of execution errors.

\subsubsection{
% Feature Consistency
Extraction Function Verification}
\label{sec:extraction_function_verification}
After using the filters to process the feature extraction functions, \workname~executes the functions that pass filtering on each raw biosignal data sample, generating the corresponding feature values. 
Specifically, for each data sample, \workname sequentially executes all feature extraction functions. 
During each function execution, \workname assigns each biosignal modality to the parameter with the corresponding modality name. 
After that, \workname verifies the outputs of each function across all samples. 
If any function produces an output with inconsistent dimensions (e.g., varying lengths, zero-length vectors) from any previous one, that function is entirely discarded and all of its previous outputs are excluded. 
Finally, the remaining valid function outputs are concatenated into a unified feature vector. 
These vectors are then stacked together and concatenated with the previous tabular feature data along the feature dimension to form updated tabular feature data.
\modify{
This verification ensures consistent feature dimensions across all samples, enabling subsequent operator-based feature generation (source 3) and feature selection on the tabular feature values.
}
% \modify{Unlike prior code verification methods that only assess execution success, this module introduces cross-sample output consistency verification, which addresses a failure mode unique to tabular feature construction: functions that execute without error but produce structurally inconsistent outputs across different samples. Such inconsistencies cannot be detected by execution-based verification alone, yet they would corrupt the integrity of the resulting tabular feature data.}
\\

\noindent \modify{Compared to most existing LLM-based code generation works that rely on execution error feedback to iteratively prompt the LLM to regenerate failing code, \workname adopts a filter-then-verify paradigm rather than a generate-debug-regenerate loop. By intercepting erroneous functions both before execution (Section~\ref{sec:filtering}) and after execution (Section~\ref{sec:extraction_function_verification}), \workname avoids returning error logs to the LLM entirely. This design choice prevents the LLM from stalling on individual failing functions across multiple rounds of correction, and more critically, avoids the rapid growth of context length that would otherwise result from accumulating error feedback in long iterative dialogues.}

\subsection{Iterative Refinement via Feature Assessment}
\begin{figure}[t]
  \centering
  \includegraphics[width=0.95\linewidth]{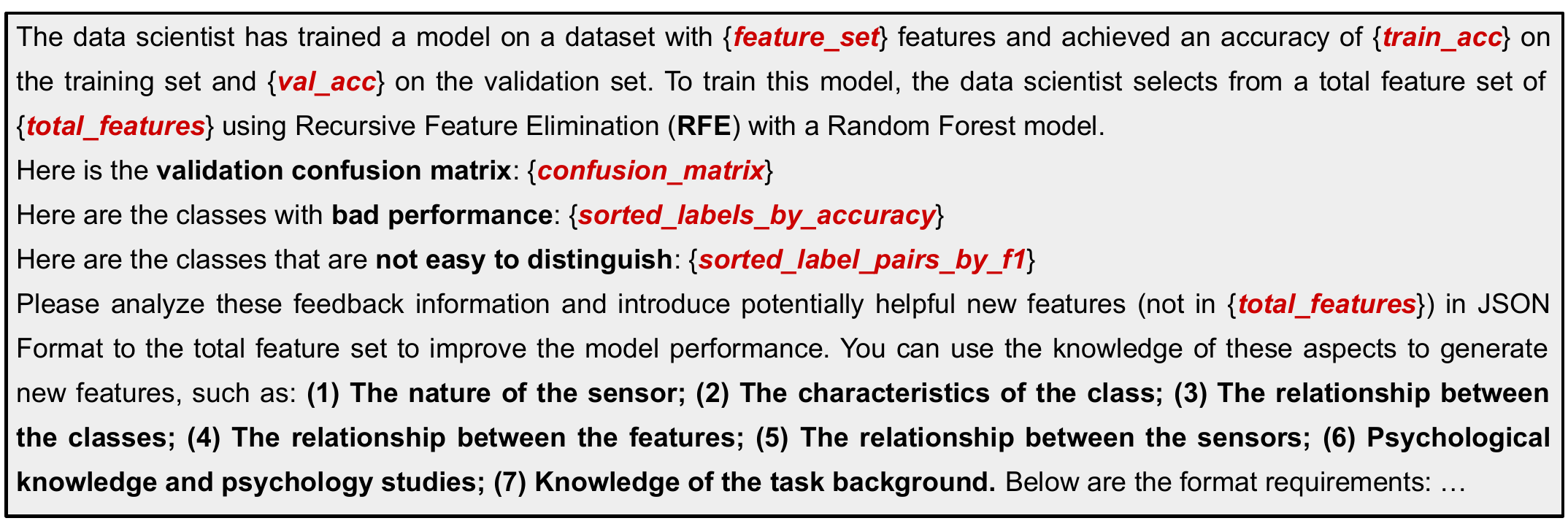}
  \caption{Feedback prompt for iterative refinement.} 
  \label{fig:feedback_prompt}
\end{figure}

\label{sec:feature_assessment}
In this section, we introduce how \workname selects features from the candidate feature set to train the downstream ML models at each iteration and constructs feedback to guide feature generation in subsequent iterations, thereby progressively refining the feature set.
\modify{
Our design goal is to leverage feature assessment results not only to select informative features but also to construct effective feedback for multi-round feature generation.
Our design details are as follows.
}

\subsubsection{Ranking-based Feature Elimination}
\label{sec:feature_selection}
At each iteration, after getting the new tabular feature data, \workname~employs a ranking-based feature selector to identify the most informative features to train the downstream models. 
This step filters out irrelevant or misleading features that may arise from the inherent randomness and hallucination in LLMs' outputs. Specifically, \workname maintains a record of the best model weights with corresponding evaluation results and selected features. 
% In each iteration, it first randomly selects a predefined proportion (e.g., 20\%) of data samples from the training set to form a validation set. 
\modify{
In each iteration, for each feature selection trial, \workname splits the original training data into a training subset (e.g., 80\%) and a validation subset (e.g., 20\%).
The splitting protocol (either sample-level or subject-level) is consistent with the protocol used to separate the final test set.
It then uses a ranking-based feature selection method, such as Recursive Feature Elimination (RFE)~\cite{guyon2002gene}, to select the required number of features from the tabular feature data. Feature selection is performed on the training subset.
The validation subset is then used to evaluate the performance of the selected features.
If the model performance in any iteration exceeds all previous models, \workname updates the best weights, results, and features.
Neither subset is used for the evaluation of the final experimental results to avoid data leakage.
To mitigate the risk of overfitting to the validation subset, the feature selector evaluates a range of target feature numbers in each iteration. 
For each target number, feature selection is repeated across multiple random seeds for the train-validation split, and the mean validation performance across seeds is used to identify the optimal feature number.
The feature subset corresponding to the best-performing seed under the optimal feature number is then retained as the selection result for that iteration.
At the end of each iteration, \workname constructs feedback to guide LLM feature generation in the next iteration.
}

\begin{algorithm}[t]
\SetKwInOut{Input}{Input}
\SetKwInOut{Output}{Output}
\SetKwFunction{FeatureSelection}{FeatureSelection}
\SetKwFunction{FeedbackConstruction}{FeedbackConstruction}
\SetKwFunction{FeatureFromLLMs}{FeatureFromLLMs}
\SetKwFunction{FeatureFromContext}{FeatureFromContext}
\SetKwFunction{FeatureFromOperators}{FeatureFromOperators}
\SetKwFunction{CodeGeneration}{CodeGeneration}
\SetKwFunction{CodeExecution}{CodeExecution}
\SetKwFunction{Concatenation}{Concatenation}
	
\Input{JSON-formatted initial feature set $S_0$, tabular feature data $X_0$ of $S_0$, feature generation stride $m$, raw biosignal data $D$, iteration number $N_{iter}$}
\Output{The model weights $W_{best}$, feature set $S_{best}$ with best performance}
\BlankLine 
\tcp{Initialization}
{Construct local task-specific contextual knowledge base}\;
$W_{best} \leftarrow None$; $S_{best} \leftarrow \emptyset$\; 
$X \leftarrow X_0$; $S \leftarrow S_0$\;
\tcp{Feature Generation Iteration}
\For{$i\leftarrow 0$ \KwTo $N_{iter}$}{
    \If{($S_{best}!=\emptyset$ or $i!=0$)} {
        $W_{temp}, S_{temp}$ $\leftarrow$ \FeatureSelection{$S$, {X}}\;
        \If{AUROC of $W_{temp}$ > AUROC of $W_{best}$} {
            $W_{best} \leftarrow W_{temp}$; $S_{best} \leftarrow S_{temp}$\;
        }
    }
    $F \leftarrow$ \FeedbackConstruction{$W_{best}$}\;
    $S_{llm} \leftarrow$ \FeatureFromLLMs{$F, S, S_{best}, m$}\;
    $S_{context} \leftarrow$ \FeatureFromContext{$F, S, S_{best}, m$}\; 
    $S \leftarrow S \cup S_{llm} \cup S_{context}$\;
    $C \leftarrow$ \CodeGeneration{$S_{llm}, S_{context}$}\;
    $X_{new} \leftarrow$ \CodeExecution{$C, D$}\;
    $X \leftarrow$ \Concatenation{$X, X_{new}$}\;
    $X, S_{op} \leftarrow$ \FeatureFromOperators{$X, S, m$}\;
    $S \leftarrow S \cup S_{op}$\;
} 
\caption{$\mathbf{ContAwaFeatGen}(S_0, X_0, m, D, N_{iter})$}
\label{algo:workflow} 
\end{algorithm}

\subsubsection{Model Performance-based Feedback}
\label{sec:results_analysis}

% After feature assessment, \workname includes the validation confusion matrix of the best model obtained from the ranking-based feature elimination module in the feedback prompt.
\modify{After feature assessment, \workname obtains the validation confusion matrix of model trained on the selected features.
However, this raw matrix is too abstract and implicit to directly guide subsequent feature generation.
To address this, \workname further derives the following more instructive information from the confusion matrix and includes them in the feedback prompt.}
\begin{itemize}
    \item \textbf{Validation accuracy:} this item aims to show the overall performance of the selected features.
    \item \textbf{Labels sorted by validation accuracies:} this item aims to highlight the classes with poor performance, guiding LLMs to generate targeted features.
    \item \textbf{Label pairs sorted by F1 scores:} This item highlights hard-to-classify classes, guiding LLMs to generate targeted features.
\end{itemize}

\noindent \workname also includes the following in the feedback prompt.
\begin{itemize}
    \item \textbf{Selected features:} this item aims to show the current effective features from the candidate feature set $S$, providing experience for LLMs to generate new features.
    \item \textbf{Candidate features:} this item aims to illustrate features that already exist to avoid duplicate generation.
\end{itemize}

\workname~uses the prompt shown in Figure~\ref{fig:feedback_prompt} to integrate all the items, which is provided for LLMs to generate new JSON-formatted features.

\subsection{Overall Algorithm}
\label{sec:overall_algorithm}
Algorithm~\ref{algo:workflow} presents the workflow of \workname.
Given system parameters, \workname initializes the best model weights $W_{best}$, best feature set $S_{best}$, current tabular data $X$, and current feature set $S$ (line 2-3).
\workname then enters an iterative feature generation process.
In each iteration, \workname first evaluates the current feature set $S$ performance and updates $W_{best}$ and $S_{best}$ accordingly (line 5-8).
Based on evaluation results, it constructs feedback information (line 9).
\workname then generates new features through two different sources: direct LLM generation (line 10) and context-guided generation (line 11). 
These new features are added to $S$ (line 12) and converted to executable code to process raw data into new columns $X_{new}$ (line 13-14).
Additional features are introduced through predefined operator combinations (line 16). This repeats until the iteration limit $N_{iter}$ is reached, then \workname outputs optimal weights $W_{best}$ and feature set $S_{best}$.

Given the vast and unbounded feature space with infinite possible feature combinations, \workname employs a heuristic search strategy to efficiently identify high-performing feature sets.
\workname tracks historically optimal features on the validation set and iteratively refines the feature selection using task-specific context and performance feedback, thereby steering the generation process toward improved outcomes.
The key advantage of our approach lies in substantially expanding the explorable feature space via the integration of three complementary sources, while minimizing manual intervention through a fully automated pipeline.

\section{EXPERIMENTAL EVALUATION}\label{exp}

In this section, we first outline the evaluation setup, which includes the datasets, tasks, and metrics (Section~\ref{sec:exp_setup}). 
We then present quantitative and qualitative results that demonstrate \workname's effectiveness and superiority over baseline methods (Section~\ref{sec:overall_performance}), followed by an ablation study examining each component's contribution (Section~\ref{sec:ablation_study}). We conduct a sensitivity analysis to assess the robustness of \workname under different parameter settings (Section~\ref{sec:sensitivity_analysis}).
Finally, we measure and analyze the system overhead of \workname (Section~\ref{sec:system_overhead}).

\subsection{Evaluation Setup}
\label{sec:exp_setup}

\subsubsection{Implementation}
We implement \workname~ using Python 3 and leverage APIs provided by VolcEngine~\cite{volcengine2024} and Microsoft Azure~\cite{azure} platforms to access LLMs. 
\modify{
To construct the Retrieval-Augmented Generation (RAG) pipeline, the source documents are first divided into chunks of up to 2,000 characters, with splitting boundaries determined by natural punctuation to preserve semantic integrity. For text embeddings, we utilize Cohere's \texttt{embed-english-v3.0} model via their APIs~\cite{cohere}. We then employ the Annoy Python package~\cite{annoy} for efficient similarity search, retrieving the top-
$k$ ($k=3$) most relevant chunks for each query. 
The RAG configuration, including chunking, embedding model, vector index, and top-$k$, is kept identical across all experiments. The knowledge base itself is task-specific: we construct and freeze one local knowledge base for each downstream task, and reuse it across random seeds and experimental settings for that task.
}
For all experiments, unless explicitly stated otherwise, we used DeepSeek-V3~\cite{liu2024deepseek} as the default LLM and Random Forest~\cite{breiman2001random} as the downstream classification model.
\modify{For all LLM-based methods, we use the same default LLM unless explicitly stated otherwise. For all feature-based methods, including \workname and the feature-library baselines, we use the same downstream Random Forest classifier and the same hyperparameter settings. Each baseline is evaluated under its native feature extraction pipeline without adding an external RFE stage in the main comparison; we further analyze the effect of applying the same RFE protocol to fixed feature-library baselines in Appendix~\ref{app:rfe_baseline_ablation}.}
The default feature generation stride was set to 20 in Section~\ref{sec:overall_performance}, and 5 in Section~\ref{sec:ablation_study} and Section~\ref{sec:sensitivity_analysis}.  
We use RFE~\cite{guyon2002gene} as the default feature selection method and the general statistical features from Section~\ref{sec:baselines} as the default initial feature set.

\subsubsection{Evaluation Datasets}\label{sec:datasets}

% As shown in Table~\ref{tab:datasets}, 
% details, good practice used 
We evaluate \workname~on four datasets covering multi-modal biosignals and diverse health tasks. The data is randomly split into 75\% training and 25\% test sets, with 20\% of the training data used for validation during feature assessment, \modify{using stratified sampling to maintain class balance.
} 
The datasets are as follows:

\noindent\textbf{SEN (Self-Collected)\footnote{\label{fn:irb}The data collection and study have been approved by the authors' IRB.}}. 
This multimodal physiological dataset includes data from 29 special education needs (SEN) students aged 3-6 years, collected using Empatica E4 wristbands~\cite{EmpaticaE4} with assistance from seven licensed teachers.
During one-on-one teaching sessions following SEN educational protocols, one teacher instructs while another observes and provides real-time emotional state annotations without interrupting the session.
The E4 wristband continuously monitors ACC, BVP, GSR, and skin temperature (ST), yielding 304 MB of data over approximately 44 hours.
We use this dataset for emotion recognition (happy, sad, neutral), which presents unique challenges where task-specific contextual knowledge is critical for generating high-quality features.
This study received institutional review board approval and followed ethical guidelines for research with vulnerable populations. Comprehensive informed consent was obtained, participation was voluntary, and children could discontinue at any time. All personally identifiable information was removed, data was anonymized with random identifiers, and secure encrypted storage was implemented with access restricted to authorized researchers.

\noindent\textbf{Epilepsy~\cite{andrzejak2001indications}}. 
This dataset contains 23.6-second single-channel EEG recordings from 10 individuals, including healthy volunteers (eyes open/closed) and epilepsy patients (seizure-free and seizure intervals).
These intracranial recordings capture activity during the seizure-free interval both within and outside the seizure-generating region, along with recordings of epileptic seizures.
In our evaluation, we employ it for two tasks: (1) epileptic seizure recognition, which identifies whether a subject is in the status of epileptic seizure.  
(2) epileptic patient recognition, distinguishing epileptic patients and healthy individuals. 

\noindent\textbf{PPG-BP~\cite{liang2018new}}. 
This dataset comprises 2.1-second PPG sequences from 219 subjects aged 20–89 years. We use it for four tasks: (1) cardiovascular disease (CVD) detection, where the goal is to identify conditions such as cerebral blood supply insufficiency or cerebrovascular disease; (2) hypertension (HTN) detection, which involves classifying individuals into Normal, Prehypertension, Stage 1, or Stage 2 Hypertension;  (3) diabetes mellitus (DM) detection, to identify whether a subject has diabetes;  and (4) cerebrovascular accident (CVA) detection, which aims to determine the presence or absence of a stroke event.

\noindent\textbf{WESAD~\cite{schmidt2018introducing}}. 
This dataset contains multi-modal physiological data collected via wrist-worn (Empatica E4~\cite{EmpaticaE4}) and chest-worn (RespiBAN~\cite{RespiBAN}) devices, including ACC, BVP, GSR, and ECG. We use it for affective state recognition, classifying participants into neutral, stress, or amusement states.
\modify{Following the data quality considerations in the NormWear ~\cite{luo2024toward}, we use these 10 selected channels as they did: \texttt{Wrist\_acc\_x/y/z}, \texttt{Wrist\_PPG}, \texttt{Wrist\_GSR}, \texttt{Chest\_acc\_x/y/z}, \texttt{Chest\_ECG}, and \texttt{Chest\_GSR}.}

For the Epilepsy, PPG-BP, and WESAD datasets, we use the initially preprocessed versions provided by the NormWear repository~\cite{normwear2024}.

\subsubsection{Baselines}
\label{sec:baselines}
We evaluate our approach against four distinct categories of solutions. 

\noindent \textbf{Expert Validated Features.} 
We compare \workname with expert-validated features. For PPG-BP dataset, we follow ~\cite{bioengineering10060678}, using heuristic-based (ABC-PSO, cuckoo clusters, dragonfly clusters) and transformation-based techniques (Hilbert transform, nonlinear regression) for dimensionality reduction. Statistical features such as mean, variance, skewness, kurtosis, PCC, and sample entropy are then computed.
For Epilepsy dataset, we follow ~\cite{Boubchir8076027} to extract time-domain, frequency-domain, and time-frequency-domain features using Quadratic Time-Frequency Distribution (QTFD).
For WESAD and SEN emotion recognition tasks, we adopt feature extraction methods from ~\cite{schmidt2018introducing} and ~\cite{liu2008physiology}, respectively, tailored to each modality.
\modify{Specifically, for WESAD, we extract the complete set of handcrafted features from Table~1 of~\cite{schmidt2018introducing}, including frequency-domain HRV components and GSR-derived features. Following~\cite{normwear2024}, we use 10 selected channels (\texttt{Wrist\_acc\_x/y/z}, \texttt{Wrist\_PPG}, \texttt{Wrist\_GSR}, \texttt{Chest\_acc\_x/y/z}, \texttt{Chest\_ECG}, and \texttt{Chest\_GSR}) based on data quality considerations. 
}

\noindent\textbf{General Statistical Features.} 
We select widely-used features from time and frequency domains. Time-domain features include mean, standard deviation, maximum, skewness, kurtosis, and quantiles (25\%, 50\%, 75\%). Frequency-domain features include spectral centroid, spread, mean frequency, peak frequency, and quantiles (25\%, 50\%, 75\%).

\noindent\textbf{Automated Feature Extraction Libraries.} 
We evaluate 4 
widely-used automated feature extraction frameworks, covering diverse methodologies from high-dimensional to low-dimensional, exhaustive to carefully curated features. 
% with both academic rigor and industrial practicality.
\begin{itemize}

% \noindent  \textbullet~ 
\item 
\textbf{TsFresh~\cite{christ2018time}}, which computes over 70 base features from raw time series using its default configuration (\texttt{Comprehensive}\-\texttt{FCParameters}). These features are expanded through operations like varying window sizes to over 794 dimensions.

% \noindent  \textbullet~ 
\item 
\textbf{TSFEL~\cite{barandas2020tsfel}}, which extracts over 65 features spanning statistical, temporal, spectral, and fractal domains.
\item 
\textbf{Kats~\cite{Kats2021}}, which generates 40 time series characteristics, including statistical properties, trend analysis, seasonality metrics, and nonlinear dynamics.
\item 
\textbf{Catch22~\cite{lubba2019catch22}}, which extracts 22 canonical features extensively validated across 93 real-world tasks.
\end{itemize}

\noindent\textbf{LLM-based Feature Engineering.} 
We compare three LLM-based automated feature engineering methods.
These baselines use context or task-specific information during LLM prompting; CAAFE and AutoIoT further adopt iterative feedback mechanisms to improve LLM-generated features.

\begin{itemize}

\item 
\textbf{CAAFE~\cite{hollmann2024large}},  which utilizes LLMs to iteratively generate new features for tabular data by using model performance improvement as feedback.

\item 
\textbf{ELLM-FT~\cite{gong2025evolutionary}}, which constructs a multi-population database using reinforcement learning and evolutionary algorithms, then incorporates selected populations into prompts to guide LLM feature generation.

\item 
\textbf{AutoIoT~\cite{shen2025autoiot}}, which autonomously gathers relevant knowledge from online sources and iteratively generates end-to-end programs for extracting sensor-data features in AIoT applications.
\end{itemize}

\modify{
\noindent\textbf{Raw-signal Modeling Baseline.}
We compare with a representative time-series classification baseline that operates directly on raw biosignals through convolutional transformations.
\begin{itemize}
\item
\textbf{MiniRocket~\cite{dempster2021minirocket}}, a fast and accurate time-series classification method based on a large set of randomized convolutional kernels. MiniRocket transforms raw time series into discriminative convolutional features and trains a lightweight linear classifier on top.
\end{itemize}
}

\noindent\textbf{End-to-end Large Multimodal Model (LMM)}.
We compare the downstream performance between models trained on features extracted using \workname with large multimodal models (LMMs) trained directly on raw biosignals. 
As our baseline, we adopt NormWear~\cite{luo2024toward}, which is presented as the first multimodal foundation model designed to learn generalized representations from wearable sensing data.
NormWear employs a tokenization module to transform biosignals into tokens, which are then processed by a share-weighted encoder. 
However, the NormWear model backbone has a size of approximately \modify{519MB}, making it impractical for long-term deployment on wearable devices due to substantial power and storage demands associated with continuous biosignal processing. 
Furthermore, unlike \workname, end-to-end models such as NormWear lack the ability to extract features with explicit semantic meaning.

\subsubsection{Evaluation Metrics}
In alignment with \workname's emphasis on healthcare applications, the Area Under the Receiver Operating Characteristic curve (AUROC) is employed as the primary evaluation metric.
AUROC provides a comprehensive measure of a model's discriminative capability across all classification thresholds. 
For multiclass tasks, AUROC is calculated using a one-vs-one approach with macro-averaging to ensure an equitable and balanced evaluation across all classes.

\begin{table*}[]
\caption{The overall performance on downstream wearable biosignal-related tasks under the sample-level split setting. \workname achieves the highest mean AUROC on PPG-BP (HTN, DM, CVA, and CVD) and SEN, while remaining competitive with the best methods, TSFEL, TsFresh, and NormWear, on Epilepsy-Disease, Epilepsy-Seizure, and WESAD, respectively. \workname significantly outperforms all baselines on PPG-BP (HTN), PPG-BP (CVD), and SEN ($p<0.05$; see Table~\ref{tab:per_task_significance_sample}).}
\label{tab:overall_performance}
\resizebox{1\linewidth}{!}{
\begin{tabular}{l|cccc|cc|cc|cc}
\hline
\toprule[1.5pt]
Methods  & \makecell[c]{PPG-BP\\ (HTN)} & \makecell[c]{PPG-BP\\ (DM)} & \makecell[c]{PPG-BP\\ (CVA)} & \makecell[c]{PPG-BP\\ (CVD)} & \makecell[c]{Epilepsy\\ (Disease)} & \makecell[c]{Epilepsy\\ (Seizure)} & WESAD & SEN & \makecell[c]{Task Avg.} & \makecell[c]{Dataset Avg.} \\  \midrule[0.7pt]
\cellcolor[HTML]{EFEFEF}Expert & 58.38$\pm$2.06  & 58.83$\pm$2.24 & 68.89$\pm$1.53  & 52.28$\pm$1.41  & 93.00$\pm$0.05   & 93.31$\pm$0.05  & 72.29$\pm$1.07  & 82.41$\pm$0.10 &  72.42$\pm$0.48 & 76.86$\pm$0.35 \\ 
\cellcolor[HTML]{EFEFEF}General & 56.67$\pm$1.12  & 50.31$\pm$1.89  & 62.17$\pm$3.91  & 57.21$\pm$1.93  & 97.78$\pm$0.03  & 97.73$\pm$0.08 & 68.55$\pm$0.49  & 89.45$\pm$0.24 & 72.48$\pm$0.61 & 78.09$\pm$0.33 \\ \hline
\cellcolor[HTML]{DAE8FC}TsFresh  & 65.29$\pm$0.61  & 50.38$\pm$3.28  & 51.92$\pm$9.09  & 56.15$\pm$3.69   & 99.58$\pm$0.01  & 99.90$\pm$0.01 & 69.79$\pm$0.48  & 86.09$\pm$0.18  & 72.39$\pm$1.30 & 77.89$\pm$0.66 \\
\cellcolor[HTML]{DAE8FC}TSFEL   & 64.66$\pm$0.79  & 68.43$\pm$1.56 & 65.27$\pm$1.26  & 54.07$\pm$5.34 & 99.67$\pm$0.02 & 99.83$\pm$0.03 & 68.88$\pm$0.14  & 87.65$\pm$0.13  & 76.06$\pm$0.72 & 79.85$\pm$0.36 \\
\cellcolor[HTML]{DAE8FC}Kats  & 65.33$\pm$0.79 & 63.05$\pm$4.31  & 54.99$\pm$3.91 & 54.75$\pm$3.99   & 99.19$\pm$0.03 & 99.51$\pm$0.02 & 71.52$\pm$0.57   & 78.03$\pm$0.08  & 73.30$\pm$0.89 & 77.11$\pm$0.47 \\
\cellcolor[HTML]{DAE8FC}Catch22  & 59.32$\pm$1.65 & 61.27$\pm$2.37 & 61.48$\pm$4.68  & 54.07$\pm$1.88 & 98.21$\pm$0.02  & 98.99$\pm$0.02 & 72.59$\pm$0.35 & 57.44$\pm$0.53 & 70.42$\pm$0.73 & 71.92$\pm$0.40 \\ \hline
\cellcolor[HTML]{FFFFC7}CAAFE  & 56.92$\pm$0.69 & 47.40$\pm$2.87 & 61.67$\pm$1.39 & 60.99$\pm$4.53  & 97.85$\pm$0.04 & 97.87$\pm$0.28 & 70.27$\pm$0.72 & 89.73$\pm$0.16  & 72.84$\pm$0.70 & 78.65$\pm$0.40 \\
\cellcolor[HTML]{FFFFC7}ELLM\_FT  & 55.26$\pm$1.81 & 52.34$\pm$0.42 & 55.55$\pm$2.20  & 53.57$\pm$1.93 & 96.55$\pm$0.74  & 96.52$\pm$0.60  & 67.73$\pm$0.65  & 87.38$\pm$0.44 & 70.61$\pm$0.46 & 76.46$\pm$0.32 \\ 
\cellcolor[HTML]{FFFFC7}AutoIoT  & 62.81$\pm$0.36 & 53.81$\pm$1.36 & 67.58$\pm$2.31  & 53.39$\pm$2.62 & 98.99$\pm$0.04  & 99.75$\pm$0.01  & 68.44$\pm$0.42  & 86.85$\pm$0.29 & 73.95$\pm$0.48 & 78.51$\pm$0.27 \\ \midrule[0.7pt]
\cellcolor[HTML]{F5DEB3}MiniRocket & 64.39$\pm$2.33 &  59.62$\pm$2.50 & 61.48$\pm$3.20 & 60.54$\pm$2.36 & 98.84$\pm$0.05 & 99.45$\pm$0.06 & 68.31$\pm$0.40 & 78.83$\pm$0.06 & 73.93$\pm$0.66 & 76.95$\pm$0.34 \\ 
\cellcolor[HTML]{F5DEB3}NormWear & 68.48$\pm$2.51 & 65.07$\pm$1.75 & 67.19$\pm$0.67  & 51.48$\pm$1.78 & 98.01$\pm$0.13 & 98.04$\pm$0.03  &  74.81$\pm$0.94 & 84.14$\pm$1.68 & 75.90$\pm$0.51 & 80.01$\pm$0.53 \\ \midrule[0.7pt]
\cellcolor[HTML]{FDDDEF}Ours & 71.97$\pm$1.44  & 69.54$\pm$2.55 & 69.73$\pm$2.51 & 70.36$\pm$2.79  & 98.44$\pm$0.27  & 99.64$\pm$0.38 & 73.20$\pm$1.70 & 92.43$\pm$0.08  & 80.66$\pm$0.63 & 83.77$\pm$0.52 \\ 
\bottomrule[1.5pt]
\end{tabular}
}
\end{table*}

\begin{table*}[]
\caption{The overall performance on downstream wearable biosignal-related tasks under the cross-subject split setting. \workname achieves the highest mean AUROC on PPG-BP (HTN, DM, and CVD), Epilepsy-Disease, and WESAD, while remaining competitive with the best methods, Expert, TsFresh, and MiniRocket, on PPG-BP (CVA), Epilepsy-Seizure, and SEN, respectively. \workname significantly outperforms all baselines on PPG-BP (DM) ($p<0.05$; see Table~\ref{tab:per_task_significance_cross_subject}).}
\label{tab:overall_performance_corss_subject}
\resizebox{1\linewidth}{!}{
\begin{tabular}{l|cccc|cc|cc|cc}
\hline
\toprule[1.5pt]
Methods  & \makecell[c]{PPG-BP\\ (HTN)} & \makecell[c]{PPG-BP\\ (DM)} & \makecell[c]{PPG-BP\\ (CVA)} & \makecell[c]{PPG-BP\\ (CVD)} & \makecell[c]{Epilepsy\\ (Disease)} & \makecell[c]{Epilepsy\\ (Seizure)} & WESAD & SEN & \makecell[c]{Task Avg.} & \makecell[c]{Dataset Avg.} \\  \midrule[0.7pt]
\cellcolor[HTML]{EFEFEF}Expert & 55.07$\pm$1.07 & 58.78$\pm$3.10 & 63.50$\pm$1.79 & 58.10$\pm$1.68 & 90.76$\pm$0.23 & 88.64$\pm$0.11 & 78.91$\pm$0.16 & 54.77$\pm$0.48 & 68.57$\pm$0.52 & 70.56$\pm$0.29 \\
\cellcolor[HTML]{EFEFEF}General & 53.45$\pm$0.92 & 48.54$\pm$1.57 & 54.03$\pm$1.08 & 51.20$\pm$5.30 & 95.81$\pm$0.05 & 93.88$\pm$0.20 & 76.54$\pm$0.64 & 52.80$\pm$0.37 & 65.78$\pm$0.72 & 69.00$\pm$0.40 \\ \hline
\cellcolor[HTML]{DAE8FC}TsFresh & 58.74$\pm$1.18 & 49.91$\pm$3.49 & 45.32$\pm$3.50 & 45.69$\pm$2.48 & 99.33$\pm$0.02 & 99.84$\pm$0.02 & 77.90$\pm$0.21 & 54.54$\pm$0.40 & 66.41$\pm$0.71 & 70.48$\pm$0.37 \\
\cellcolor[HTML]{DAE8FC}TSFEL & 58.19$\pm$0.60 & 63.50$\pm$3.74 & 40.75$\pm$3.37 & 48.56$\pm$0.87 & 99.21$\pm$0.01 & 99.67$\pm$0.01 & 76.87$\pm$0.55 & 54.73$\pm$0.26 & 67.69$\pm$0.65 & 70.95$\pm$0.36 \\
\cellcolor[HTML]{DAE8FC}Kats & 57.08$\pm$1.82 & 61.63$\pm$3.15 & 40.49$\pm$2.00 & 51.40$\pm$3.19 & 98.72$\pm$0.04 & 98.82$\pm$0.18 & 77.61$\pm$0.40 & 55.95$\pm$0.13 & 67.71$\pm$0.66 & 71.25$\pm$0.34 \\
\cellcolor[HTML]{DAE8FC}Catch22 & 56.48$\pm$1.55 & 57.12$\pm$1.65 & 40.34$\pm$1.69 & 56.48$\pm$2.22 & 97.57$\pm$0.02 & 98.67$\pm$0.11 & 75.65$\pm$0.10 & 53.22$\pm$0.04 & 66.94$\pm$0.45 & 69.90$\pm$0.23 \\ \hline
\cellcolor[HTML]{FFFFC7}CAAFE & 52.91$\pm$0.72 & 51.88$\pm$3.27 & 56.86$\pm$3.09 & 52.10$\pm$4.37 & 95.84$\pm$0.02 & 93.69$\pm$0.34 & 78.50$\pm$0.28 & 52.90$\pm$0.47 & 66.83$\pm$0.79 & 69.90$\pm$0.42 \\
\cellcolor[HTML]{FFFFC7}ELLM\_FT & 51.36$\pm$3.28 & 51.92$\pm$1.87 & 60.41$\pm$3.17 & 49.12$\pm$7.60 & 94.21$\pm$0.26 & 88.40$\pm$0.47 & 75.83$\pm$0.56 & 52.37$\pm$0.81 & 65.45$\pm$1.14 & 68.18$\pm$0.62 \\
\cellcolor[HTML]{FFFFC7}AutoIoT & 62.42$\pm$0.91 & 50.42$\pm$1.19 & 59.02$\pm$8.55 & 50.43$\pm$6.76 & 98.45$\pm$0.05 & 99.23$\pm$0.05 & 78.38$\pm$0.05 & 55.11$\pm$0.39 & 69.18$\pm$1.37 & 71.98$\pm$0.69 \\ \midrule[0.7pt]
\cellcolor[HTML]{F5DEB3}MiniRocket & 66.01$\pm$0.92 & 57.63$\pm$3.91 & 51.38$\pm$6.27 & 52.16$\pm$4.57 & 98.05$\pm$0.05 & 97.75$\pm$0.12 & 76.92$\pm$0.48 &  56.97$\pm$0.44 & 69.61$\pm$1.10 & 72.15$\pm$0.57 \\ 
\cellcolor[HTML]{F5DEB3}NormWear & 61.32$\pm$4.54 & 56.66$\pm$7.99 & 56.10$\pm$2.99 & 57.35$\pm$6.50 & 94.59$\pm$2.96 & 96.51$\pm$0.51 & 67.74$\pm$0.85 & 52.93$\pm$1.70 & 67.90$\pm$1.52 & 68.52$\pm$0.95 \\ \midrule[0.7pt]
\cellcolor[HTML]{FDDDEF}Ours & 66.17$\pm$3.41 & 66.38$\pm$4.16 & 62.45$\pm$1.01 & 65.80$\pm$0.23 & 99.56$\pm$0.15 & 98.70$\pm$0.33 & 80.09$\pm$0.97 & 54.62$\pm$1.56 & 74.22$\pm$0.72 & 74.76$\pm$0.57 \\
\bottomrule[1.5pt]
\end{tabular}
}
\end{table*}

\subsection{Overall Performance}
\label{sec:overall_performance}

\subsubsection{Quantitative Results}
\modify{
To evaluate the performance of \workname, we employ two different data split protocols: (1) Sample-level split: all samples are randomly mixed and divided into training and testing sets (Table~\ref{tab:overall_performance}). (2) Subject-level split: subjects are randomly partitioned into training and testing sets, ensuring no subject overlap between the two sets (Table~\ref{tab:overall_performance_corss_subject}).
\final{We report two average AUROC metrics in both tables. The task-weighted average is computed by directly averaging AUROC across the eight downstream tasks, so each task contributes equally. The dataset-balanced average first averages task AUROCs within each dataset and then averages across datasets, so each dataset contributes equally regardless of the number of tasks it contains.}
We use three random seeds and repeat the experiments for each method on each dataset under these seeds, reporting the mean AUROC and standard deviation.
\final{
Table~\ref{tab:overall_performance} and Table~\ref{tab:overall_performance_corss_subject} show that \workname achieves the highest task-weighted and dataset-balanced average AUROC in both settings. For the task-weighted average, \workname outperforms the best baselines by 4.60\% and 4.61\% under the sample-level and subject-level settings, respectively; for the dataset-balanced average, it outperforms the best baselines by 3.76\% and 2.61\%, respectively.}
\final{Under the sample-level split, \workname achieves the highest task-weighted and dataset-balanced average AUROC of 80.66\% and 83.77\%, respectively, with the largest gains on the four PPG-BP tasks and SEN. Under the subject-level split, \workname again achieves the highest task-weighted and dataset-balanced average AUROC of 74.22\% and 74.76\%, respectively, with the most pronounced improvements on the PPG-BP tasks, top-performing results on Epilepsy-Disease and WESAD, and competitive performance on Epilepsy-Seizure and SEN.}

\noindent\textbf{The magnitude of performance improvement varies across different tasks.} \workname exhibits substantial performance improvement on tasks including PPG-BP HTN, DM, CVA, and CVD, while achieving comparable performance to the best baseline on Epilepsy-Disease, Epilepsy-Seizure, and SEN.
% This is because the datasets for PPG-BP HTN, DM, CVA, and CVD are relatively small, and the performance of each task is highly correlated with its contextual characteristics, making it more sensitive to feature selection. 
% This further highlights the greater advantages of \workname for wearable biosignal-driven tasks where data collection and annotation are more difficult.
% In contrast, the remaining tasks benefit from larger datasets or lower task difficulties, where sufficient sample size partially compensates for performance variations across different features. As a result, these tasks are less sensitive to feature choices and thus offer relatively limited room for further improvement.
% \workname can achieve the best or comparable results on various tasks with distinct characteristics, demonstrating greater generality than other baselines.
This is because the PPG-BP dataset is relatively smaller in scale (PPG-BP: 657, SEN: 26227, Epilepsy: 11500, WESAD: 11050). 
With small dataset scale, downstream models lack sufficient samples to learn discriminative patterns.
Consequently, the model performance becomes highly dependent on the quality of the selected features.
\workname can generate a broader set of high-quality, task-specific features, leading to a substantial performance lead across all four tasks in PPG-BP.
This is corroborated by Tables~\ref{tab:overall_performance} and Table~\ref{tab:overall_performance_corss_subject}, where the larger baseline variance on PPG-BP indicates high sensitivity of small-scale tasks to feature selection results.
% These observations emphasize that the key advantage of our method lies in its universality.
\final{These observations emphasize that the key advantage of our method lies in its adaptability across heterogeneous tasks.}
Because the ``best-performing baseline'' fluctuates significantly depending on the specific task, picking a single optimal baseline for every scenario is infeasible; instead, \workname utilizes a single, unified framework to stably achieve top-tier results comparable to the state-of-the-art across a wide variety of tasks, demonstrating its superior adaptability and robustness.

\noindent \textbf{Limitations of each type of baselines.}
Expert-validated features, though competitive, fail to achieve the best results because these features are not perfectly matched to our task context. In contrast, \workname can dynamically and precisely customize feature sets for highly context-specific tasks, rather than relying on static, generalized expert knowledge.
In addition, different from the baselines with a relatively fixed number of features, such as ``Expert'', ``General'', and automated feature extraction libraries (e.g., TsFresh, TSFEL, Kats, Catch22), \workname can iteratively generate and introduce new features into the candidate space, continuously pushing the performance boundaries.
Compared to the first two LLM-based methods, CAAFE and ELLM-FT, which rely on operations over existing features, \workname integrates three different feature sources. This enables \workname to have much larger candidate feature space and outperform these methods by up to 10.05\% (Table~\ref{tab:overall_performance}) and 8.77\% (Table~\ref{tab:overall_performance_corss_subject}).
AutoIoT~\cite{shen2025autoiot}, the most relevant LLM-based baseline, suffers from over 90\% erroneous feature extraction functions, causing its detailed logs in the auto-correction mechanism to exceed LLM token limits. 
For a fair comparison, we manually selected only the correctly functioning feature extraction functions, but \workname still outperformed AutoIoT by a significant margin.
Although AutoIoT also includes a debug module to automatically handle code execution errors, which is similar to the role of our multi-layer execution function filtering and execution verification modules, it struggles to conduct feature generation iteration as smoothly as \workname.
This is because 
(1) We categorize potential error types and implement corresponding filters for progressive filtering, rather than directly sending the lengthy exception logs to the LLM.
This enables \workname~to identify potential errors more accurately and efficiently.
(2) \workname~validates both function execution and feature extraction outputs, filtering additional erroneous code. 
(3) \workname~adopts a filtering-and-discarding strategy instead of iterative debugging until the code becomes correct, ensuring smooth iteration while preventing massive token accumulation.
\modify{Simply chunking these exception logs to fit token limits is insufficient, as it remains difficult to precisely identify useful information from the complex and diverse error messages.}
Compared to end-to-end deep learning baselines (e.g., MiniRocket and NormWear), our massive feature space better assists the model in extracting comprehensive information from limited samples.}

\begin{figure}[t]
  \centering
  \includegraphics[width=\linewidth]{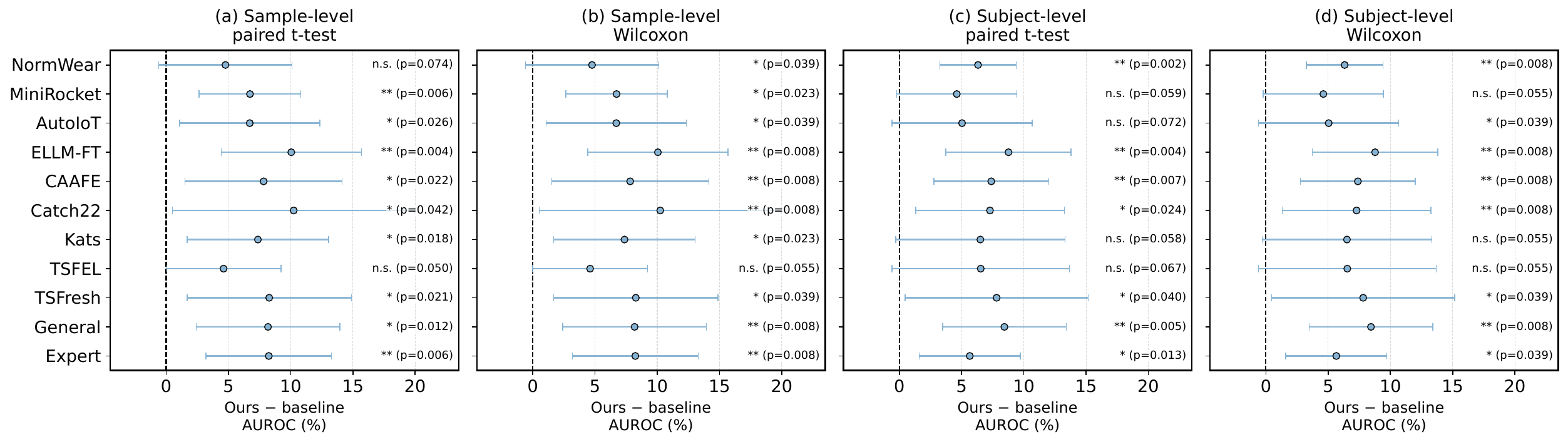}
  \caption{\modify{Statistical significance analysis of \workname against baseline methods under sample-level and subject-level split settings.
Each point denotes the mean paired AUROC improvement of \workname over a baseline across the eight downstream tasks, and horizontal error bars indicate the 95\% confidence intervals of the paired task-level differences.
The vertical dashed line indicates zero improvement.
Panels (a) and (c) report paired t-test results, while panels (b) and (d) report Wilcoxon signed-rank test results.
Significance levels and p-values are annotated next to each method.}}
  \label{fig:significance_analysis}
\end{figure}

\modify{\subsubsection{Statistical Significance Analysis}
Figure~\ref{fig:significance_analysis} summarizes the statistical significance analysis under both sample-level and subject-level split settings.
For each baseline, we compute the paired AUROC improvement of \workname over the baseline across the eight downstream tasks, and report the mean improvement with 95\% confidence intervals.
For each task, the AUROC used in the paired test is the mean over three independent random seeds under the corresponding split protocol; therefore, the paired samples in this analysis are the downstream tasks. Under the sample-level split setting, both the paired t-test and the Wilcoxon signed-rank test indicate statistically significant improvements over most individual baselines.
A similar trend is observed under the more challenging cross-subject split setting.
\workname maintains positive average improvements over most individual baselines, showing that its advantage is not limited to sample-level random splits but also generalizes to subject-level evaluation.
}

\begin{figure}[]
    \centering
    \begin{minipage}{0.47\linewidth}
        \includegraphics[width=\linewidth]{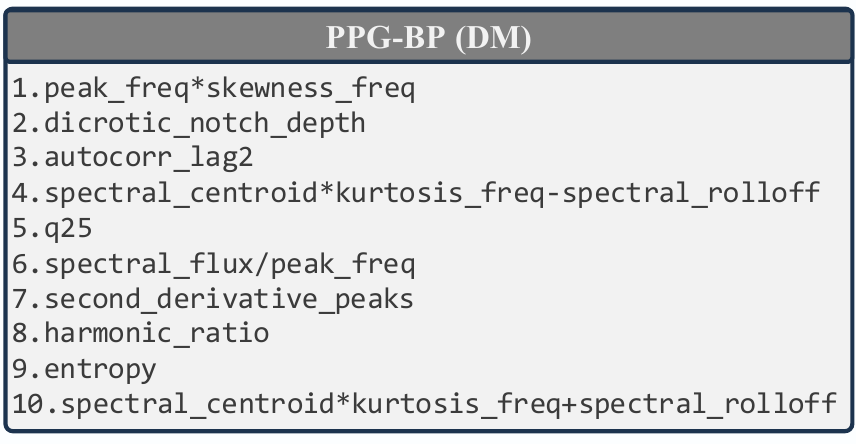}
         \vspace{-8mm}
        \label{fig:features_ppg_bp_dm}
    \end{minipage}
    \hspace{1mm}
    \begin{minipage}{0.47\linewidth}
        \includegraphics[width=\linewidth]{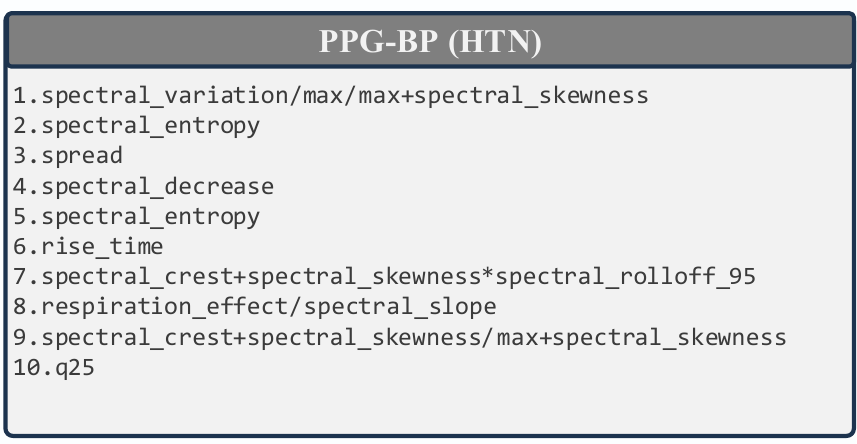}
        \vspace{-8mm}
        \label{fig:features_ppg_bp_htn}
    \end{minipage}
    \\[10pt]
    \begin{minipage}{0.47\linewidth}
        \includegraphics[width=\linewidth]{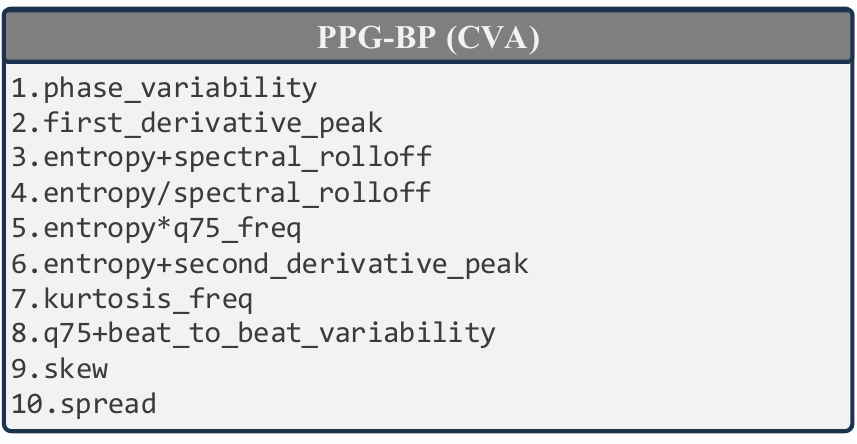}
         \vspace{-8mm}
        \label{fig:features_ppg_bp_cva}
    \end{minipage}
    \hspace{1mm}
    \begin{minipage}{0.47\linewidth}
        \includegraphics[width=\linewidth]{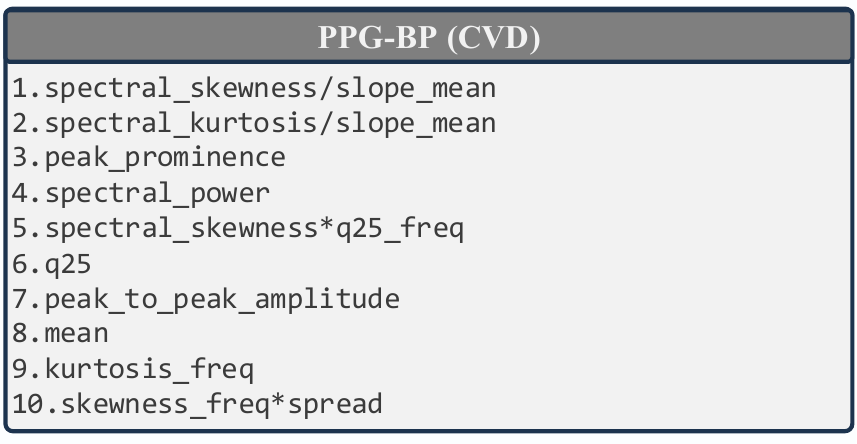}
        \vspace{-8mm}
        \label{fig:features_ppg_bp_cvd}
    \end{minipage}
    \\[10pt]
    \begin{minipage}{0.47\linewidth}
        \includegraphics[width=\linewidth]{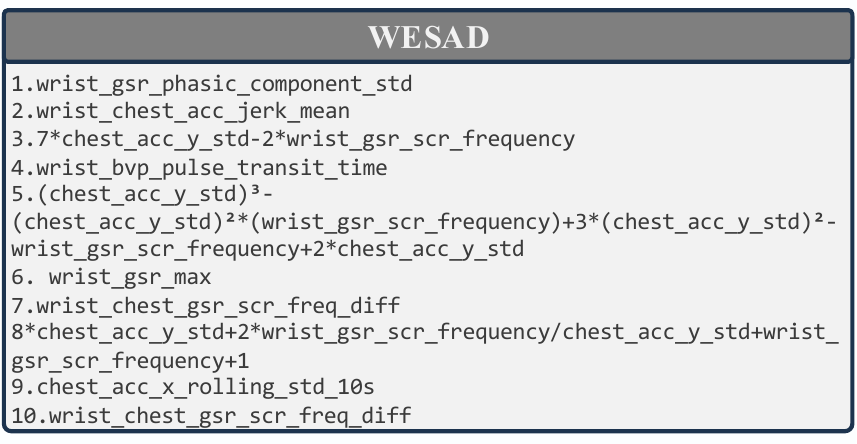}
         \vspace{-8mm}
        \label{fig:features_wesad}
    \end{minipage}
    \hspace{1mm}
    \begin{minipage}{0.47\linewidth}
        \includegraphics[width=\linewidth]{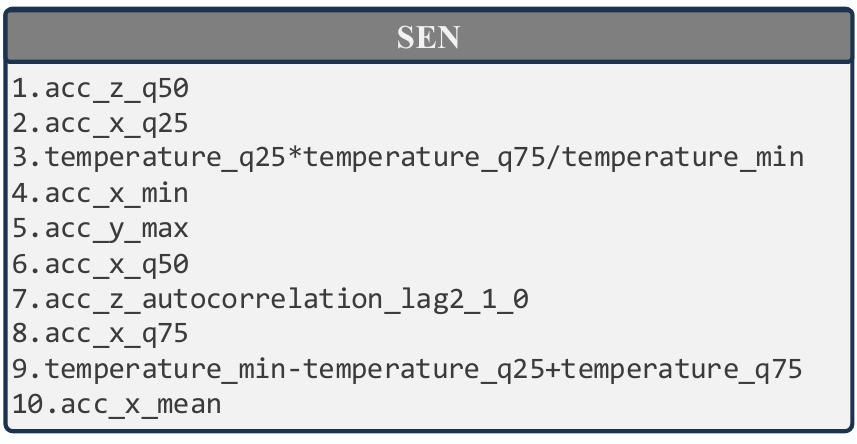}
        \vspace{-8mm}
        \label{fig:features_sen}
    \end{minipage}
    \\[10pt]
    \begin{minipage}{0.47\linewidth}
        \includegraphics[width=\linewidth]{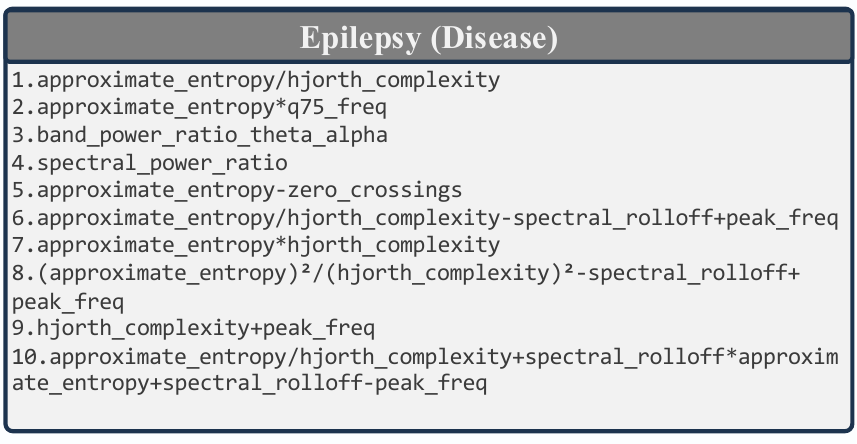}
         \vspace{-5mm}
        \label{fig:features_epilepsy_disease}
    \end{minipage}
    \hspace{1mm}
    \begin{minipage}{0.47\linewidth}
        \includegraphics[width=\linewidth]{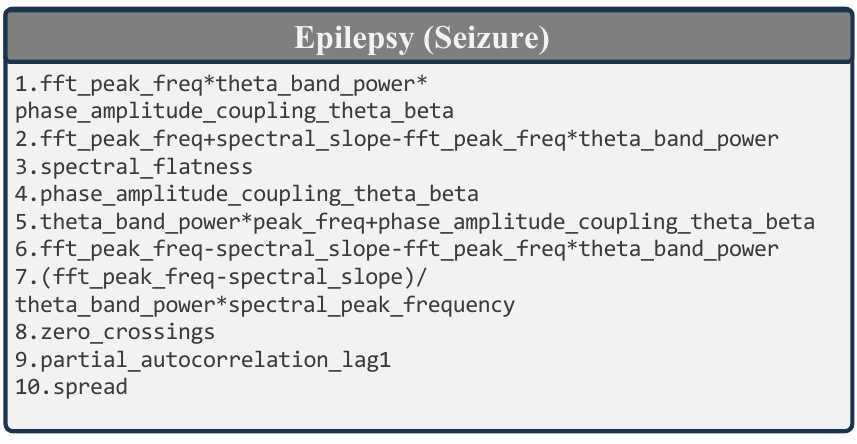}
        \vspace{-5mm}
        \label{fig:features_epilepsy_seizure}
    \end{minipage}
    \caption{Top-10 features identified for each task.}
    \label{fig:qualitative_results}
\end{figure}

% 我们系统能handle很多任务，泛化能力，能给很多任务赋能

% 使用chunking不work，因为单轮的已经超了
% 很难log里面找到有用的信息

\begin{figure}[t]
  \centering
  % \begin{minipage}[t]{0.46\linewidth}
    \centering
    \includegraphics[width=0.5\linewidth]{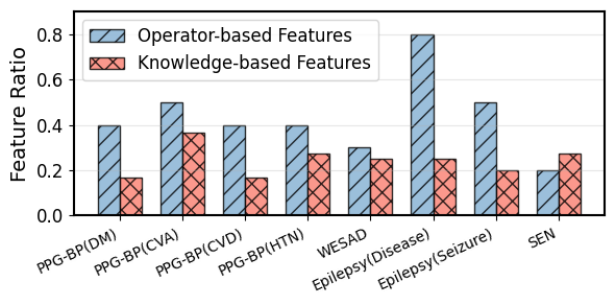}
    \caption{Feature Source Distribution Analysis. We examine the origins of the top-10 important features. The blue bars indicate the proportion of features derived through operator-based transformations, while the orange bars show the proportion of features involving task-specific contextual knowledge.}
    \label{fig:feature_ratio}
\end{figure}

\subsubsection{Qualitative Results}

We analyze representative cases from each task to demonstrate \workname's effectiveness. For each case, Figure~\ref{fig:qualitative_results} presents the top-10 most influential features ranked by RFE importance scores. We trace each feature's source and show the source proportions in Figure~\ref{fig:feature_ratio}. Key observations are as follows:

\noindent\textbf{\textit{Observation 1: The most important features vary significantly across different tasks.}} 
As shown in Figure~\ref{fig:qualitative_results}, the feature generation and final selection show obvious differences across the eight tasks. 
% For example, in emotion recognition, typical individuals rely heavily on autonomic nervous system signals (GSR, heart rate, chest acceleration) strongly linked to arousal, while SEN children depend more on movement patterns and temperature changes, likely due to less stable autonomic responses and greater reliance on external behaviors or sensory sensitivities. 
% TODO: references 
This demonstrates that high-performance models require a task-specific feature set rather than a fixed, universal set. 
The optimal feature combinations vary substantially across tasks, reflecting the inherent heterogeneity in task characteristics. 
This adaptability highlights \workname's key advantage: unlike traditional handcrafted selection or automated extraction libraries, it dynamically identifies and selects more customized features for each specific task.

\noindent\textbf{\textit{Observation 2: Contextual knowledge plays a pivotal role in effective feature generation.}} 
Figure~\ref{fig:feature_ratio} shows that the top-10 important features for every task include features derived from task-specific knowledge.
For example, among the top-10 features selected for the PPG-BP (DM) task shown in Figure~\ref{fig:qualitative_results}, 4 features either wholly or partially stem from the second feature generation source: ``\texttt{peak\_freq*skewness\_freq}'', ``\texttt{dicrotic\_notch\_\\depth}'', ``\texttt{spectral\_flux\slash peak\_freq}'' and ``\texttt{harmonic\_ratio}''. This finding demonstrates that contextual knowledge effectively complements and enhances the inherent capabilities of LLMs.

\noindent\textbf{\textit{Observation 3: Operator-generated combined features demonstrate high importance.}} 
Figure~\ref{fig:feature_ratio} shows that the top-10 features of all tasks contain high proportions of features generated by operators.
The proportion of operator-generated features ranges from 20\% to 80\% across the eight tasks. This significant presence of combined features highlights the role of inter-feature interactions in achieving high-performance downstream models, validating the effectiveness of operator-based feature generation source.

\subsection{Ablation Study}
\label{sec:ablation_study}

\begin{figure}[t]
  \centering
  \includegraphics[width=0.95\linewidth]{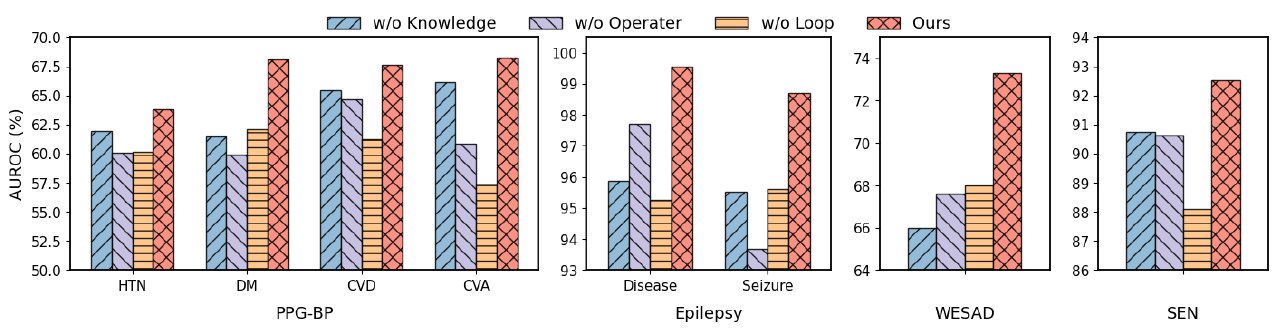}
  \caption{\modify{Ablation Study of \workname. ``w/o'' denotes the exclusion of a specific component of \workname.}}
  \label{fig:ablation_study}
\end{figure}

To validate each component's contribution, we conduct an ablation study on \modify{all the 8 tasks}.
We analyze three components: 
(1) contextual knowledge-enhanced feature generation, 
(2) operator-based feature composition, and 
(3) the iterative feedback loop. 
Each component is evaluated by removing it while keeping other settings constant.
For the feedback loop, we compare \workname~with generating the same number of features in a single iteration.
Figure~\ref{fig:ablation_study} shows that \workname~achieves the highest AUROC across all tasks, demonstrating the importance of its components.
Removing contextual knowledge decreases AUROC by \modify{3.60}\% on average, as feature generation becomes overly reliant on the LLM's inherent knowledge, limiting feature diversity and quality.
Removing operator-based composition reduces AUROC by \modify{5.52}\%, highlighting its importance in capturing intra-feature and cross-sensor relationships.
Excluding the iterative feedback loop decreases AUROC by \modify{4.60}\%, as it eliminates the adaptive refinement enabled by evaluation feedback.
These results demonstrate that each component contributes to \workname's robustness and ability to generate high-quality features across diverse tasks.

\subsection{Sensitivity Analysis}
\label{sec:sensitivity_analysis}

This section examines how system parameter settings affect \workname's performance on the PPG-BP (HTN) dataset, including the number of features introduced per iteration, the initial feature sets, feature selection methods, the choice of LLMs, and the downstream models.

\begin{figure*}[t]
    \centering
    \begin{minipage}{0.33\linewidth}
    \centering
        \includegraphics[width=\linewidth]{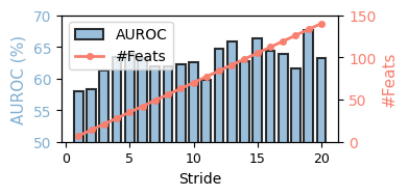}
        \vspace{-7mm}
        \subcaption{Different iteration stride.}
        \label{fig:impact_of_stride}
    \end{minipage}
    \hfill
    \begin{minipage}{0.33\linewidth}
    \centering
        \includegraphics[width=0.8\linewidth]{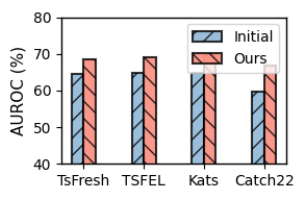}
        \vspace{-4mm}
        \subcaption{Different initial feature sets.}
        \label{fig:impact_of_initial_feature_set}
    \end{minipage}
    \hfill
    \begin{minipage}{0.33\linewidth}
    \centering
        \includegraphics[width=0.8\linewidth]{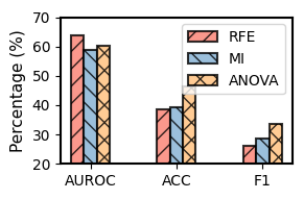}
        \vspace{-4mm}
        \subcaption{Different feature selection methods.}
        \label{fig:impact_of_feature_selection}
    \end{minipage}
    \hfill
    \begin{minipage}{0.45\linewidth}
    \centering
        \includegraphics[width=0.95\linewidth]{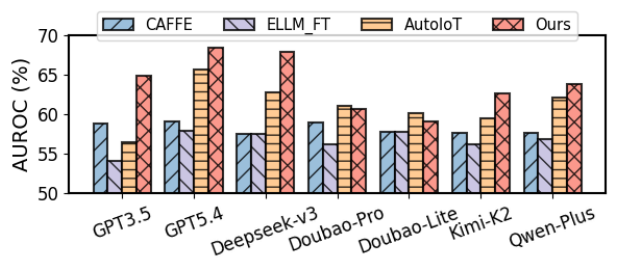}
        \vspace{-3mm}        \subcaption{\modify{Different LLMs.}}
        \label{fig:impact_of_different_LLMs}
    \end{minipage}
    \hspace{5mm}
    \begin{minipage}{0.50\linewidth}
    \centering
        \includegraphics[width=\linewidth]{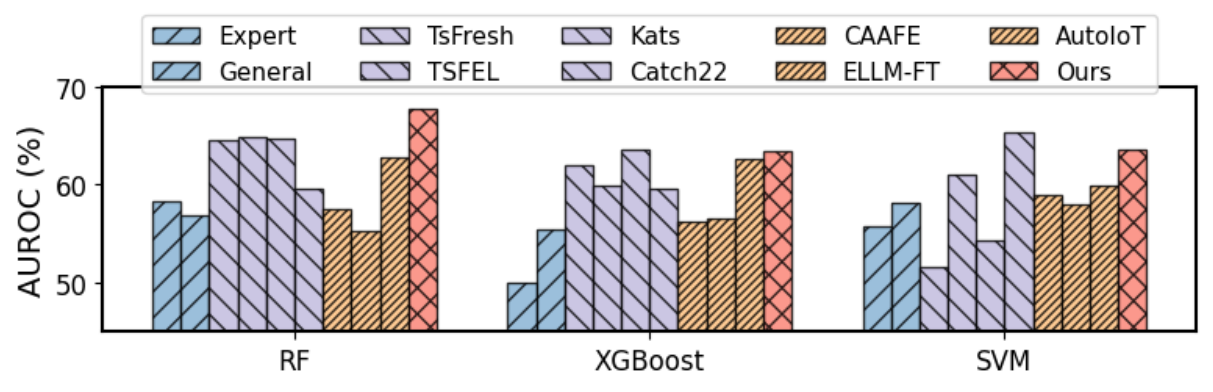}
        \vspace{-6mm}
        \subcaption{Different downstream models.}
        \label{fig:impact_of_different_downstream_models}
    \end{minipage}
    \caption{Performance of \workname under different settings on PPG-BP (HTN) dataset.}
    \label{fig:featminer_extended_performance}
\end{figure*}

\subsubsection{Impact of Iteration Stride}
As shown in Fig. \ref{fig:featminer_extended_performance}(\subref{fig:impact_of_stride}), we vary the iteration stride from 1 to 20 to examine how the number of features introduced per iteration affects downstream model performance.
For a given stride $n$, the maximum number of features added per iteration is $7n$, comprising $n$ LLM-generated features, $2n$ knowledge-augmented features (from arXiv~\cite{arxiv} and PubMed~\cite{pubmed}), and $4n$ recombinative features produced by operators.
Due to feature filtering and error handling, the actual retained features are typically fewer than the theoretical maximum. 
In this experiment, each of the 20 trials is run for exactly 10 iterations. 
According to Figure~\ref{fig:impact_of_stride}, the final AUROC exhibits an overall upward trend as the stride increases. 
This is caused by two reasons: 
1) Larger strides generate more diverse features per iteration, improving the likelihood of discovering effective ones, whereas small strides produce more homogeneous features.
2) Larger strides supply more candidate functions, increasing the probability of retaining more features after filtering and error handling.
With small strides, each iteration may yield too few or even zero executable feature-extraction functions. 
However, trials with strides 11, 14, 18, and 20 deviate from this trend due to the inherent randomness of LLM outputs, which yielded fewer useful features and reduced performance.

\subsubsection{Impact of Initial Feature Set}

Figure \ref{fig:featminer_extended_performance}(\subref{fig:impact_of_initial_feature_set}) compares the performance of the TsFresh~\cite{christ2018time}, TSFEL~\cite{barandas2020tsfel}, Kats~\cite{Kats2021}, and Catch22~\cite{lubba2019catch22} before and after using their features as \workname's initial feature sets, respectively. 
The results show that \workname~significantly improves the performance of all initial feature sets, with the highest AUROC increase reaching 7.16\%. 
Moreover, the performance gap between different initial feature sets narrows from 5.26\% to 2.85\%. 
This demonstrates that \workname yields substantial gains regardless of the starting feature quality and lifts all sets to a comparable performance level.

\subsubsection{Impact of Feature Selection Method}
As shown in Figure~\ref{fig:featminer_extended_performance}(\subref{fig:impact_of_feature_selection}), we investigate the impact of non-LLM feature selection methods on the performance of the downstream model. 
In this experiment, we use a stride of 5 and run 10 iterations, evaluating the \workname~on the PPG-BP HTN task. 
We compare the performance of three classical feature selection methods: 
1) Recursive Feature Elimination (RFE)~\cite{guyon2002gene} with Random Forest as the base model. 
2) Mutual Information (MI)~\cite{lewis1992feature}. 
3) Analysis of Variance (ANOVA)~\cite{fisher1970statistical}. Figure~\ref{fig:impact_of_feature_selection} indicates that \workname~performs best with RFE. 
This is because the RFE with a nonlinear base model has superior capabilities in nonlinear detection, interaction capture, and noise resistance compared to MI and ANOVA~\cite{diaz2006gene, guyon2002gene}. 
However, its computational intensity reduces efficiency. The choice of feature selection methods requires balancing system efficiency with model performance.

\subsubsection{Impact of Different LLMs}
\label{sec:sensitivity_analysis_llms}
We compare the performance of \workname with three baseline LLM-based methods across 7 different LLMs, including DeepSeek-V3 ~\cite{liu2024deepseek}, GPT-3.5~\cite{openai_chatgpt_2024}, \modify{GPT-5.4~\cite{openai_gpt54_2026}}, Doubao-Pro~\cite{doubao_pro_2024}, Doubao-Lite~\cite{doubaolite128k}, Kimi-K2~\cite{team2025kimi} and Qwen-Plus~\cite{team2024qwen2}. 
As shown in Fig.~\ref{fig:featminer_extended_performance}(\subref{fig:impact_of_different_LLMs}), \workname exhibits consistent and robust performance across all evaluated LLMs, outperforming the baseline methods in most cases. 
GPT-5.4 achieves the highest performance, showcasing its advanced capabilities. 
Doubao-Lite, while weaker than advanced models, still outperforms LLM-based baselines.

\subsubsection{Impact of Different Downstream Models}
\label{sec:sensitivity_analysis_downstream_model}
Figure~\ref{fig:featminer_extended_performance}(\subref{fig:impact_of_different_downstream_models}) shows the impact of the choice of downstream model on the performance. We compare our approach with the baselines using three different types of machine learning models: Random Forest (RF)~\cite{breiman2001random}, XGBoost~\cite{chen2016xgboost}, and Support vector machine (SVM)~\cite{hearst1998support}. 
% We set the stride to be 5 and the total iteration number to be 10. 
The results show that \workname~can achieve the highest AUROC or comparable performance to the best-performing baseline using most of the downstream models.

\subsection{System Overhead}
\label{sec:system_overhead}

\begin{figure}[t]
  \centering
  \includegraphics[width=0.5\linewidth]{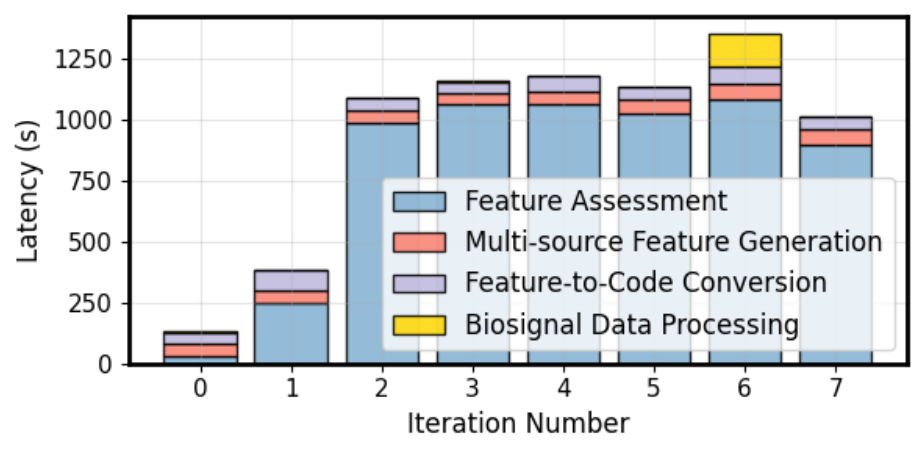}
  \caption{Cumulative latency consumption of different stages across iterations.}
  \label{fig:system_overhead}
\end{figure}

Figure~\ref{fig:system_overhead} shows the latencies per iteration of the four stages in \workname across iterations using the PPG-BP HTN task (stride=10, iterations=10).
Feature assessment dominates total latency, accounting for approximately 85\% of the time, with variability depending on dataset size, feature selection methods, and selection strategies. 
In this experiment, we adopted the RFE method and evaluated multiple target feature numbers, which contributes to the higher latency.
The remaining 15\% of the total overhead (approximately 20 minutes) is attributed to LLM API calls, code generation, and raw biosignal processing.

This latency represents a strategic trade-off, where the investment in offline computation is justified by the significant performance gains achieved in the final model. 
As \workname is designed for offline feature engineering, this overhead is acceptable for many healthcare applications where the primary goal is to obtain a highly performant and robust model. 
The resulting lightweight feature set is then ideal for efficient, long-term deployment on resource-constrained wearable devices. 
Future work will focus on improving efficiency through the use of lightweight LLMs, faster feature selection algorithms, and parallelization of feature function execution.

\section{DISCUSSION}
\noindent \textbf{Scalability to Other Data Signals.} 
Currently, \workname~focuses on biosignal feature generation and already demonstrates versatility across multiple data types. 
\modify{Future work will explore the \workname's performance on diverse time-series data beyond biosignals, such as industrial sensors, and human activity recognition.
For non-time-series data such as depth camera images and radar point clouds, which are used in many healthcare applications,}
\workname~successfully extracts meaningful statistical features such as mean and variance of pixel intensity. However, higher-order, learning-based features still require trainable models rather than handcrafted functions. 
Future work could extend \workname~by incorporating pretrained backbone networks to extract learning-based features from heterogeneous sensor data.
% explore the capability of DF in other time series signals

\noindent \textbf{System Efficiency.}
\workname~focuses on improving model performance, but each iteration is time-consuming, with time mainly spent on feature assessment, LLM API calls, and raw biosignal processing.
\modify{However, we observe that ML models for most downstream tasks approach their final performance after only 4-5 iterations of feature generation, demonstrating that our framework achieves strong results without requiring extensive iteration time.}
Future work will improve efficiency by using lightweight LLMs, implementing faster feature selection algorithms, and parallelizing  the execution of feature functions to reduce latency.
% how many features are selected
% how many iterations are usually required

\noindent \textbf{System Stability.}
Due to LLM stochasticity, \workname~occasionally generates features that hurt downstream performance.
We will explore experience-recording mechanisms to capture common patterns in successful features to guide future generation, and conduct more rigorous cross-validation before applying features.
We will also try to incorporate human-in-the-loop mechanisms into \workname, leveraging human feedback to improve stability.

\noindent \textbf{Correctness and \final{Logical Validity} of Generated Code.}
Our current system focuses primarily on the downstream model's final performance, ensuring feature syntactic correctness and utility through multi-layer filtering, post-execution validation, and feature selectors.
\final{These checks help remove invalid, incomplete, or uninformative feature implementations, but they do not exhaustively validate the physiological, behavioral, or causal logic of every LLM-generated feature.
In future work, we plan to strengthen logical validation by developing more systematic automated semantic validation methods for generated features and improving code-level validation through validation agents and automated test case generation. We will also explore human-in-the-loop mechanisms to incorporate domain-expert review for assessing the physiological or clinical plausibility of generated features.}

\section{RELATED WORK}\label{relatedwork}

Extracting features from wearable biosensor data is critical for understanding physiological states and enabling interventions. Feature engineering approaches fall into three categories: handcrafted, automated, and LLM-based generation.

\noindent \textbf{Handcrafted Features.}
Handcrafted feature engineering relies on domain expertise to design task-specific features derived from time- and frequency-domain analyses. Time-domain features include mean, variance, and signal energy, while frequency-domain features leverage FFT and DWT to compute power spectral density (PSD) and other characteristics~\cite{bagirathan2021recognition}. Handcrafted features have been widely applied, from stress detection~\cite{smets2018large} using GSR, ECG, and ST to disease classification tasks such as stroke prediction~\cite{hussain2021big} and arrhythmia detection~\cite{azariadi2016ecg}. While effective, this approach is labor-intensive and lacks scalability.

\noindent \textbf{Automated Feature Engineering without LLMs.}
Automated methods reduce manual effort by generating features directly from raw sensor data. Tools like TsFresh~\cite{christ2018time}, TSFEL~\cite{barandas2020tsfel}, Kats~\cite{Kats2021}, and Catch22~\cite{lubba2019catch22} extract predefined feature sets using statistical, spectral, and decomposition techniques. However, these pipelines are typically rigid and fail to adapt to task-specific needs, limiting their flexibility in various wearable biosignal applications.
For tabular data, other automated methods include expansion-reduction approaches~\cite{kanter2015deep, katz2016explorekit, khurana2016cognito, horn2019autofeat, zhang2023openfe}, evolutionary algorithms~\cite{khurana2018feature, zhu2022evolutionary, ying2023self}, and AutoML-based pipelines~\cite{chen2019neural, bahri2022automl, wang2021autods, zhu2022difer}. 
However, these methods are limited to mathematical transformations of existing features and cannot leverage raw sensor data to extract meaningful, task-relevant features, further highlighting the need for more adaptive approaches.

\noindent\textbf{LLM-based Feature Engineering.}
LLMs have recently been investigated for feature engineering in tabular data~\cite{jeong2024llm, hollmann2024large, nam2024optimized, gong2025evolutionary}, leveraging domain-specific knowledge to generate or select useful features. 
Studies~\cite{jeong2024llm, choi2022lmpriors, li2025exploring} have demonstrated that prompting LLMs with task descriptions enables them to outperform conventional methods in feature selection by using contextual knowledge.
Frameworks such as CAAFE~\cite{hollmann2024large}, SMARTFEAT~\cite{lin2023smartfeat}, and OCTree~\cite{nam2024optimized} incorporate optimization and reasoning to iteratively improve feature quality, while LFG~\cite{zhang2024dynamic} employs agent-based strategies to refine generated features.
However, these studies focus on tabular data and cannot generate features directly from raw biosignals. 

Recent works also leverage (multi-modal) LLMs directly to understand sensor data. Unfortunately, the high inference costs and latency of LLMs make them unsuitable for long-term edge deployment with continuous biosignal collection. In addition, LLMs are typically trained for general-purpose tasks, making them difficult to apply to niche tasks and data modalities.
\cite{gao2024leveraging} leverages LLMs for generating mobile sensing strategies in human behavior modeling, but it generates features and strategies in a single forward pass, without an iterative refinement mechanism that could adaptively improve the performance of downstream models based on feedback.
AutoIoT~\cite{shen2025autoiot}, the most relevant work, can automatically synthesize code for AIoT applications. However, when applied to biosignal feature extraction, it often generates erroneous feature extraction functions, and its code improvement module frequently exceeds the token limits of LLMs.

\section{CONCLUSION}
We introduce \workname, the first LLM-empowered, context-aware feature generation framework specifically designed for wearable biosignals. 
We propose a multi-source feature generation mechanism to provide comprehensive task-specific context to LLMs, enabling tailored feature generation.
We also develop an iterative feature generation strategy and feature-to-code translation approach with in-loop feedback, enabling LLMs to adaptively refine the generated feature set over time. 
% \modify{Experiments show \workname achieves the highest average AUROC under both sample-level and subject-level evaluation settings across diverse tasks, outperforming the best baselines by 4.56\% and 4.61\%, respectively, demonstrating the promise of \workname for enabling wearable biosignal-based healthcare applications.}
\final{We extensively evaluate \workname on one self-collected and three publicly available datasets covering eight healthcare tasks. The experimental results demonstrate that \workname achieves the highest average AUROC under both sample-level and subject-level settings, with particularly pronounced gains on PPG-BP tasks and competitive performance on Epilepsy, WESAD, and SEN datasets.}
\final{
\section*{Acknowledgments}
This work was supported in part by the Research Grants Council of Hong Kong under Grant Nos. STG1/E-403/24-N and 14201425, and the Hong Kong Applied Science and Technology Research Institute under Innovation and Technology Fund Project PRP/012/22CI.
}
\section*{Ethics Statement and AI Usage}

This work does not raise any ethical issues. Claude Sonnet 4.5 was used to improve the quality of writing, including style, phrasing, and grammar in this paper. The used prompt is: \textit{Please help me polish the following text to make it more professional and academic:} [text].

\bibliographystyle{ACM-Reference-Format}
\bibliography{literature}
\clearpage
\appendix
\section{APPENDIX}
\subsection{Prompt Details}

\modify{
In this section, we provide the prompts used in \workname.
}

\modify{
Figure~\ref{fig:prompt_source_1} presents the prompt template for direct feature generation by LLMs (source 1).
The system prompt is composed of two parts: the general description of the task and the feedback prompt template.
Specifically, the template contains several placeholders designed for dynamic replacement: \texttt{task\_description} for the general task description, \texttt{feature\_set} for the previously selected features, \texttt{train\_acc} for the training accuracy of the downstream model,  \texttt{val\_acc} for the validation accuracy of the downstream model, \texttt{total\_features} for the entire candidate feature set, \texttt{confusion\_matrix} for the confusion matrix obtained on the validation subset, \texttt{sorted\_labels\_accuracy} for a list of labels sorted by the validation accuracy, \texttt{sorted\_label\_pairs\_by\_f1} for a list of label pairs sorted by their pairwise F1 scores, and \texttt{new\_feature\_stride} for the number of new features to be introduced in each iteration.
Some additional format requirements are provided in the user prompt.
}

\modify{
Figure~\ref{fig:prompt_source_2} shows the prompt template for task context-guided feature generation (source 2).
This feature source contains two steps: keywords generation and feature generation.
The system prompt for keyword generation includes the detailed task settings, \texttt{task\_setting}, and feedback prompt template.
The \texttt{new\_keyword\_stride} slot is designed to be replaced by the required number of new keywords introduced in each iteration.
The remaining slots have the same functions as the corresponding slots in Figure~\ref{fig:prompt_source_1}.
The user prompt for keyword generation provides additional format requirements of the generated keywords.
Upon receiving the keywords from the LLMs, \workname appends this response as a message with the role of \texttt{assistant}.
Then the user prompt for feature generation is appended.
The \texttt{retrieved\_knowledge} slot is designed to be replaced by the expert knowledge retrieved using the generated keywords.
}

\modify{
Figure~\ref{fig:feature_to_code_translation} presents the prompt template used in the feature-to-code translation module.
The \texttt{task\_description} slot is replaced with the general task description.
\texttt{feature\_to\_convert} slot is replaced with the feature description in JSON list format.
\texttt{sensor\_names} slot is replaced with the sensor names list used in the current task, helping emphasize function naming conventions to the large language model.
This template also provides two examples of the feature extraction functions.
The user prompt provides additional information about the data types of each function's input arguments.
}

\subsection{RFE Ablation on Baselines}
\label{app:rfe_baseline_ablation}

\modify{
Our experiments in Section 4 evaluate each baseline under its native and standard pipeline. 
We do not impose an additional RFE step on every baseline for the following reasons.
% The reasons are as follows:
First, for some baselines, such as CAAFE~\cite{hollmann2024large}, ELLM\_FT~\cite{gong2025evolutionary}, and AutoIoT~\cite{shen2025autoiot}, the final number of generated features is dynamically determined by the method itself and can vary substantially across runs. In such cases, it is neither easy nor methodologically meaningful for them to fix a uniform number of selected features in each run.
Second, some baselines (e.g., ELLM\_FT~\cite{gong2025evolutionary}) already include their own feature selection, filtering, or compression mechanisms as part of the method. We therefore treat these components as part of the original baseline pipeline. Applying an additional, external feature selection step on top of them could alter the intended workflow of the method and make the evaluation less faithful to its native setting.
Third, the feature number produced by different baselines differs substantially. Imposing an additional unified RFE step on all methods may override the design choices that these baselines make regarding feature construction and retention. As a result, such a procedure may not reflect the true native performance of the baselines.
}

\begin{figure}[H]
  \centering
  \includegraphics[width=0.70\linewidth]{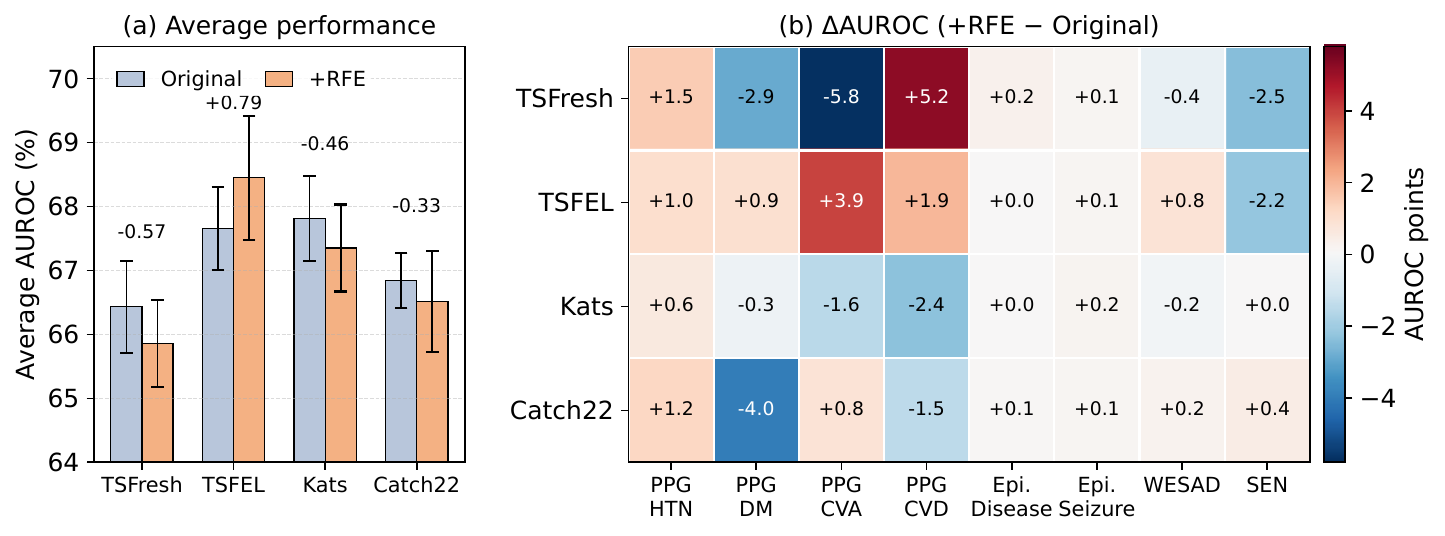}
  \caption{
  \modify{
RFE ablation on fixed feature-library baselines.
(a) Average AUROC before and after RFE.
(b) Task-wise AUROC change after RFE.
RFE does not consistently improve baseline performance, indicating that \workname's gain is not simply due to feature selection.
}
}
  \label{fig:rfe_baseline_ablation}
\end{figure}

\modify{
For \workname, the RFE-based selection step is part of the system design that prevents uncontrolled feature growth and identifies the most informative subset for downstream prediction. 
Thus, RFE is an integral part of our framework rather than an external post-processing advantage.
}
\modify{
Nonetheless, to further support our claim, we additionally applied the same RFE-based feature-selection pipeline used in our method to fixed feature-library baselines, including TSFresh, TSFEL, Kats, and Catch22, as shown in Figure~\ref{fig:rfe_baseline_ablation}.
We keep all hyperparameters of the feature selector identical to those used in our method, including the candidate target feature numbers, the 80/20 stratified train-validation split protocol, the number of random seeds, the Random Forest estimator, and the validation criterion.
The results in Figure~\ref{fig:rfe_baseline_ablation} show that applying the same RFE pipeline does not systematically improve these baselines.
The average AUROC changes from 66.43 to 65.86 for TSFresh, from 67.66 to 68.45 for TSFEL, from 67.81 to 67.35 for Kats, and from 66.84 to 66.51 for Catch22. These changes are small and task-dependent rather than consistently beneficial. Therefore, the superior performance of our method cannot be attributed merely to the use of the RFE selection pipeline; instead, it mainly comes from the quality of the generated features.
}

\final{
\subsection{Leave-One-Subject-Out Evaluation on WESAD and SEN}

To further assess subject-level generalization on these two datasets, we conduct an additional leave-one-subject-out (LOSO) evaluation on WESAD and SEN.
In each LOSO fold, one subject is held out for testing and the model is trained on the remaining subjects, so that every subject is evaluated exactly once. 
This protocol avoids the dependence on a small number of random subject splits and provides a more exhaustive estimate of cross-subject robustness.
The LOSO results in Table~\ref{tab:loso_wesad_sen} show that \workname achieves the highest mean AUROC on both datasets, reaching 80.09 on WESAD and 55.28 
on SEN. On WESAD, \workname slightly outperforms NormWear and Kats, which are the strongest baselines under this protocol. 
On SEN, \workname also obtains the best mean AUROC, with a small margin over MiniRocket. 
These results indicate that \workname remains competitive and robust under a more exhaustive subject-level evaluation setting, rather than relying on a small number of random subject splits.
}

\subsection{Per-task Statistical Significance.}
Tables~\ref{tab:per_task_significance_sample} and \ref{tab:per_task_significance_cross_subject} report the per-task statistical comparisons between \workname and each baseline using two-sided paired $t$-tests over the same three random seeds used in Table~\ref{tab:overall_performance} and~\ref{tab:overall_performance_corss_subject}.
Under the sample-level split, \workname significantly outperforms the baselines in 67 out of 88 task--baseline comparisons (76.14\%).
Under the cross-subject split, it significantly outperforms the baselines in 58 out of 88 comparisons (65.91\%).
At the task level, \workname significantly outperforms all 11 baselines on PPG-BP (HTN), PPG-BP (CVD), and SEN under the sample-level split, and on PPG-BP (DM) under the cross-subject split.
The relative performance and ranking of the baseline methods vary substantially across downstream tasks.
Despite this task-dependent variation, \workname significantly outperforms a majority of the baselines in 13 of the 16 task--split settings and remains competitive with the strongest baseline methods in several of the remaining settings.
This consistent performance across heterogeneous tasks and evaluation protocols demonstrates the strong task adaptability and robustness of \workname.

\begin{promptbox}[Prompt Template for Source 1]
\textbf{SYSTEM PROMPT:}\\
You are an assistant of a data scientist who is working on a machine learning model of a task. 
Here is the task description: [{task\_description}]

You have trained a model on a dataset with [{feature\_set}] features and achieved an accuracy of [{train\_acc}] on the training set and [{val\_acc}] on the validation set. To train this model, the data scientist selects from a total feature set of [{total\_features}] using Recursive Feature Elimination (RFE) with a Random Forest model.

Here is the validation confusion matrix: [{confusion\_matrix}]\\
Here are the classes with bad performance: [{sorted\_labels\_by\_accuracy}]\\
Here are the classes that are not easy to distinguish: [{sorted\_label\_pairs\_by\_f1}]

Please analyze this feedback information and introduce potentially helpful new features (not in [{total\_features}]) in JSON format to the total feature set to improve the model performance. \\
You can use the knowledge of the many aspects to generate new features, such as: 

(1) The nature of the sensor; \\
(2) The characteristics of the class; \\
(3) The relationship between the classes; \\
(4) The relationship between the features; \\
(5) The relationship between the sensors; \\
(6) Psychological knowledge and psychology studies; \\
(7) Knowledge of the task background. 

Please provide a list of JSON objects, each containing the following fields:\\
(1) "feature\_name": the name of the new feature\\
(2) "description": a brief description of the new feature\\
(3) "reason": why this feature could be useful. You may refer to the results shown above.

Please provide [{new\_feature\_stride}] effective new features.

\textbf{USER PROMPT:}

Now please generate some new features in the form of a JSON list to improve the model performance. \\
Follow the format provided above. \\
DO NOT generate ANY other text except for the features list. \\
If the model performance information is not provided, please give the features based on the task description and the sensor data characteristics.
\end{promptbox}
\captionof{figure}{\modify{Prompt template for feature generation in source 1.}}
\label{fig:prompt_source_1}

\begin{promptbox}[Prompt Template for Source 2]
\textbf{SYSTEM PROMPT (KEYWORD GENERATION):}\\
You are an assistant of a data scientist who is working on a machine learning model of a task. 
Here are some important details about this task: [{task\_setting}]

You have trained a model on a dataset with [{feature\_set}] features and achieved an accuracy of [{train\_acc}] on the training set and [{val\_acc}] on the validation set. To train this model, the data scientist selects from a total feature set of [{total\_features}] using Recursive Feature Elimination (RFE) with a Random Forest model.

Here is the validation confusion matrix: [{confusion\_matrix}]\\
Here are the classes with bad performance: [{sorted\_labels\_by\_accuracy}]\\
Here are the classes that are not easy to distinguish: [{sorted\_label\_pairs\_by\_f1}]

Please analyze this information and introduce some keywords that can be used to retrieve potentially helpful knowledge from the digital library. The retrieved knowledge will be used as references to generate new features. You can use the knowledge of many aspects to generate new keywords, such as: 

(1) The nature of the sensor; \\
(2) The characteristics of the class; \\
(3) The relationship between the classes; \\
(4) The relationship between the features; \\
(5) The relationship between the sensors; \\
(6) Psychological knowledge and psychology studies; \\
(7) Knowledge of the task background. 

Please provide a list of strings, each one is a keyword that can be used to retrieve relevant knowledge from the digital library.\\
Please provide [{new\_keyword\_stride}] most promising keywords based on the analysis of current model performance.

\textbf{USER PROMPT (KEYWORD GENERATION):}\\
Now please give the keywords in the form of a string list to improve the model performance. \\
Follow the format provided above. \\
DO NOT generate ANY other text except the keywords list. \\
If the model performance information is not provided, please give the keywords based on the task settings.

\textbf{ASSISTANT PROMPT (FEATURE GENERATION):}\\
- [{keywords}].

\textbf{USER PROMPT (FEATURE GENERATION):}\\
Now, we have retrieved some knowledge from the digital library using the keywords you provided: [{keywords}]. \\
Please generate some new features to improve the model performance based on the retrieved knowledge: [{retrieved\_knowledge}].

The generated features should follow the following format:
These features should be organized as a list of JSON objects, each containing the following fields: \\
(1) "feature\_name": the name of the new feature \\
(2) "description": a brief description of the new feature \\
(3) "reason": why this feature could be useful. You may refer to the results shown above. 
    
Please provide [{new\_feature\_stride}] most effective new features in the form of a JSON list. DO NOT generate ANY other text except for the features list.

\end{promptbox}
\captionof{figure}{\modify{Prompt template for feature generation in source 2.}}
\label{fig:prompt_source_2}

\begin{promptbox}[Prompt Template for Feature-to-Code Translation]
\textbf{SYSTEM PROMPT:} \\
You are an assistant of a data scientist who is working on a machine learning model of a task. 
Here is the task description: [{task\_description}]

Here are the features to be extracted from the sensor data. Please convert them into Python code: [{features\_to\_convert}]\\

Each feature is a Python function that takes the sensor data as input and returns the feature values in a dictionary with the feature names as keys and the feature values as values. \\
If the feature requires multiple sensor data, you can pass them as separate arguments. 

If the feature is applicable to a specific sensor data, the function name should start with the sensor name based on its inputs (for the current task, the function name must start with the following sensor names as prefix: [{sensor\_names}]. e.g., "gsr\_XXX", "wrist\_gsr\_XXX". You must choose from this list.) \\
The parameters of the function corresponding to the sensor data should also be named as the sensor name mentioned above.

Pay attention to the data characteristics and the feature calculation method.\\
Here are some examples of the functions:

Example 1:
\begin{verbatim}
def gsr\_mean(gsr):
    mean_data = np.mean(gsr)
    return {
        "mean_gsr": mean_data
    }
\end{verbatim}

Example 2:
\begin{verbatim}
def acc_multi_correlation(acc_x, acc_y, acc_z, onset_threshold=0.2):
    data = np.stack([acc_x, acc_y, acc_z], axis=1)
        
    displacement = np.sqrt(np.sum(data**2, axis=1))
    velocity = np.diff(data, axis=0)  
    acceleration = np.diff(velocity, axis=0) 
        
    displacement = displacement[2:]
        
    displacement_thresh = np.max(displacement) * onset_threshold
    accel_magnitude = np.sqrt(np.sum(acceleration**2, axis=1))
    accel_thresh = np.max(accel_magnitude) * onset_threshold
        
    onset_idx = np.where((displacement > displacement_thresh) & 
        (accel_magnitude > accel_thresh))[0][0]
        
    temporal_features = {
        'reaction_time': onset_idx + 2, 
        'time_to_peak_velocity': np.argmax(np.sqrt(np.sum(velocity**2, axis=1))) + 1,  
        'time_to_peak_acceleration': np.argmax(accel_magnitude) + 2  
    }
        
    for key, value in temporal_features.items():
        if isinstance(value, np.generic):
            temporal_features[key] = value.item()
        
    return temporal_features
\end{verbatim}

Directly provide the converted Python code. Do not include any explanation or other text.

\textbf{USER PROMPT:} \\
Assume that each signal is provided as a one-dimensional native Python LIST to the feature extraction function. For example:
\begin{verbatim}
    ppg = [x_1, x_2, ...]
\end{verbatim}
Now please convert the features into Python code. 
\end{promptbox}
\captionof{figure}{\modify{Prompt template for feature-to-code translation.}}
\label{fig:feature_to_code_translation}

\begin{table}[H]
\centering
\caption{The performance on WESAD and SEN under the leave-one-subject-out split setting.}
\label{tab:loso_wesad_sen}
\resizebox{1\linewidth}{!}{
\begin{tabular}{l|cc|cccc|ccc|cc|c}
\hline
\toprule[1.5pt]
Dataset
& \cellcolor[HTML]{EFEFEF}Expert
& \cellcolor[HTML]{EFEFEF}General
& \cellcolor[HTML]{DAE8FC}TsFresh
& \cellcolor[HTML]{DAE8FC}TSFEL
& \cellcolor[HTML]{DAE8FC}Kats
& \cellcolor[HTML]{DAE8FC}Catch22
& \cellcolor[HTML]{FFFFC7}CAAFE
& \cellcolor[HTML]{FFFFC7}ELLM\_FT
& \cellcolor[HTML]{FFFFC7}AutoIoT
& \cellcolor[HTML]{F5DEB3}MiniRocket
& \cellcolor[HTML]{F5DEB3}NormWear
& \cellcolor[HTML]{FDDDEF}Ours \\
\midrule[0.7pt]
WESAD LOSO & 76.44$\pm$8.06 & 72.93$\pm$6.17 & 79.58$\pm$6.24 & 78.70$\pm$6.42 & 79.72$\pm$6.21 & 76.75$\pm$5.98 & 74.74$\pm$5.78 & 71.70$\pm$6.21 & 78.17$\pm$6.63 & 77.98$\pm$6.40 & \underline{79.78$\pm$6.74} & \textbf{80.09$\pm$5.40} \\
SEN LOSO & 52.38$\pm$10.24 & 51.01$\pm$12.00 & 53.40$\pm$13.22 & 51.78$\pm$12.73 & 54.43$\pm$11.21 & 54.42$\pm$11.04 & 51.98$\pm$7.87 & 52.59$\pm$11.52 & 53.34$\pm$11.81 & \underline{55.22$\pm$10.66} & 53.22$\pm$9.48 & \textbf{55.28$\pm$10.65} \\
\bottomrule[1.5pt]
\end{tabular}
}
\end{table}

\begin{table}[H]
\caption{Per-task significance between \workname and each baseline under the sample-level split setting, computed using the results from three random seeds used in Table~\ref{tab:overall_performance}. Cells report two-sided paired $t$-test $p$-values.}
\label{tab:per_task_significance_sample}
\centering
\resizebox{1\linewidth}{!}{
\begin{tabular}{l|cccc|cc|cc}
\toprule[1.5pt]
Methods & \makecell[c]{PPG-BP (HTN)} & \makecell[c]{PPG-BP (DM)} & \makecell[c]{PPG-BP (CVA)} & \makecell[c]{PPG-BP (CVD)} & \makecell[c]{Epilepsy (Disease)} & \makecell[c]{Epilepsy (Seizure)} & WESAD & SEN \\
\midrule[0.7pt]
\cellcolor[HTML]{EFEFEF}Expert & ${0.0100}$ & ${0.0070}$ & $0.6985$ & ${0.0105}$ & ${0.0011}$ & ${0.0014}$ & $0.5959$ & ${1.33\times 10^{-6}}$ \\
\cellcolor[HTML]{EFEFEF}General & ${0.0092}$ & ${0.0173}$ & ${0.0303}$ & ${0.0399}$ & ${0.0467}$ & ${0.0180}$ & ${0.0345}$ & ${0.0010}$ \\ \hline
\cellcolor[HTML]{DAE8FC}TsFresh & ${0.0188}$ & ${0.0226}$ & $0.0939$ & ${0.0424}$ & ${0.0181}$ & $0.3669$ & $0.1137$ & ${8.29\times 10^{-5}}$ \\
\cellcolor[HTML]{DAE8FC}TSFEL & ${0.0297}$ & $0.2119$ & ${0.0452}$ & ${0.0476}$ & ${0.0146}$ & $0.5005$ & ${0.0409}$ & ${3.65\times 10^{-5}}$ \\
\cellcolor[HTML]{DAE8FC}Kats & ${0.0290}$ & ${0.0236}$ & ${0.0267}$ & ${0.0168}$ & ${0.0343}$ & $0.5992$ & $0.1333$ & ${<10^{-6}}$ \\
\cellcolor[HTML]{DAE8FC}Catch22 & ${0.0005}$ & ${0.0014}$ & ${0.0303}$ & ${0.0137}$ & $0.2552$ & $0.0975$ & $0.5328$ & ${5.51\times 10^{-5}}$ \\ \hline
\cellcolor[HTML]{FFFFC7}CAAFE & ${0.0061}$ & ${0.0072}$ & ${0.0446}$ & ${0.0126}$ & ${0.0471}$ & ${0.0407}$ & ${0.0355}$ & ${0.0003}$ \\
\cellcolor[HTML]{FFFFC7}ELLM\_FT & ${0.0002}$ & ${0.0078}$ & ${0.0007}$ & ${0.0062}$ & ${0.0326}$ & ${0.0272}$ & ${0.0233}$ & ${0.0017}$ \\
\cellcolor[HTML]{FFFFC7}AutoIoT & ${0.0048}$ & ${0.0067}$ & ${0.0101}$ & ${0.0257}$ & $0.0628$ & $0.6640$ & ${0.0408}$ & ${0.0005}$ \\ \hline
\cellcolor[HTML]{F5DEB3}MiniRocket & ${0.0046}$ & ${8.47\times 10^{-6}}$ & ${0.0023}$ & ${0.0006}$ & $0.0878$ & $0.4119$ & ${0.0228}$ & ${7.21\times 10^{-7}}$ \\
\cellcolor[HTML]{F5DEB3}NormWear & ${0.0299}$ & ${0.0105}$ & $0.1393$ & ${0.0010}$ & ${0.0336}$ & ${0.0156}$ & $0.0679$ & ${0.0122}$ \\
\bottomrule[1.5pt]
\end{tabular}
}
\end{table}

\begin{table}[H]
\caption{Per-task significance between \workname and each baseline under the subject-level split setting, computed using the results from three random seeds used in Table~\ref{tab:overall_performance_corss_subject}. Cells report two-sided paired $t$-test $p$-values.}
\label{tab:per_task_significance_cross_subject}
\centering
\resizebox{1\linewidth}{!}{
\begin{tabular}{l|cccc|cc|cc}
\toprule[1.5pt]
Methods & \makecell[c]{PPG-BP (HTN)} & \makecell[c]{PPG-BP (DM)} & \makecell[c]{PPG-BP (CVA)} & \makecell[c]{PPG-BP (CVD)} & \makecell[c]{Epilepsy (Disease)} & \makecell[c]{Epilepsy (Seizure)} & WESAD & SEN \\
\midrule[0.7pt]
\cellcolor[HTML]{EFEFEF}Expert & ${0.0303}$ & ${0.0149}$ & $0.1450$ & ${0.0157}$ & ${0.0005}$ & ${0.0006}$ & $0.1717$ & $0.9093$ \\
\cellcolor[HTML]{EFEFEF}General & ${0.0356}$ & ${0.0221}$ & ${0.0029}$ & ${0.0379}$ & ${0.0007}$ & ${0.0025}$ & ${0.0291}$ & $0.2427$ \\ \hline
\cellcolor[HTML]{DAE8FC}TsFresh & ${0.0287}$ & ${0.0048}$ & ${0.0223}$ & ${0.0056}$ & $0.1003$ & ${0.0250}$ & ${0.0403}$ & $0.9360$ \\
\cellcolor[HTML]{DAE8FC}TSFEL & ${0.0400}$ & ${0.0396}$ & ${0.0098}$ & ${0.0012}$ & $0.0593$ & ${0.0343}$ & ${0.0136}$ & $0.8957$ \\
\cellcolor[HTML]{DAE8FC}Kats & ${0.0165}$ & ${0.0345}$ & ${0.0032}$ & ${0.0174}$ & ${0.0084}$ & $0.7155$ & ${0.0184}$ & $0.2823$ \\
\cellcolor[HTML]{DAE8FC}Catch22 & ${0.0326}$ & ${0.0296}$ & ${0.0031}$ & ${0.0215}$ & ${0.0014}$ & $0.9077$ & ${0.0160}$ & $0.2517$ \\ \hline
\cellcolor[HTML]{FFFFC7}CAAFE & ${0.0135}$ & ${0.0244}$ & ${0.0432}$ & ${0.0294}$ & ${0.0007}$ & ${4.37\times 10^{-6}}$ & $0.0712$ & $0.2224$ \\
\cellcolor[HTML]{FFFFC7}ELLM\_FT & ${0.0449}$ & ${0.0089}$ & $0.2435$ & $0.0599$ & ${0.0007}$ & ${0.0009}$ & ${0.0314}$ & ${0.0371}$ \\
\cellcolor[HTML]{FFFFC7}AutoIoT & $0.2419$ & ${0.0354}$ & $0.5133$ & $0.0552$ & ${0.0049}$ & $0.1276$ & $0.0925$ & $0.6204$ \\ \hline
\cellcolor[HTML]{F5DEB3}MiniRocket & $0.9215$ & ${0.0003}$ & $0.0677$ & ${0.0321}$ & ${0.0015}$ & ${0.0159}$ & $0.0632$ & $0.0681$ \\
\cellcolor[HTML]{F5DEB3}NormWear & ${0.0176}$ & ${0.0481}$ & ${0.0309}$ & $0.1447$ & $0.0921$ & ${0.0022}$ & ${3.15\times 10^{-5}}$ & ${0.0023}$ \\
\bottomrule[1.5pt]
\end{tabular}
}
\end{table}

\end{document}